\documentclass[journal]{IEEEtran}
\usepackage{amsmath,amsfonts}
\usepackage{algorithmic}
\usepackage{algorithm}
\usepackage{booktabs}
\usepackage{array}
\usepackage{multirow}
\usepackage[caption=false,font=normalsize,labelfont=sf,textfont=sf]{subfig}
\usepackage{textcomp}
\usepackage{stfloats}
\usepackage{url}
\usepackage{verbatim}
\usepackage{graphicx}
\usepackage{cite}
\usepackage{hyperref}
\usepackage{float}
\usepackage{makecell}
\usepackage[table]{xcolor}
\definecolor{crimson}{rgb}{0.86, 0.08, 0.24}
\usepackage[switch]{lineno}

\begin{document}
\title{Learning Where, What and How to Transfer: A Multi-Role Reinforcement Learning Approach for Evolutionary Multitasking}

\author{Jiajun Zhan, Zeyuan Ma,  Yue-Jiao Gong,~\IEEEmembership{Senior Member,~IEEE} and  Kay Chen TAN,~\IEEEmembership{Fellow,~IEEE} 
\thanks{Corresponding author: Yue-Jiao Gong~(E-mail: gongyuejiao@gmail.com)}
}



\maketitle

\begin{abstract}
Evolutionary multitasking~(EMT) algorithms typically require tailored designs for knowledge transfer, in order to assure convergence and optimality in multitask optimization. In this paper, we explore designing a systematic and generalizable knowledge transfer policy through Reinforcement Learning. We first identify three major challenges: determining the task to transfer~(where), the knowledge to be transferred~(what) and the mechanism for the transfer (how). To address these challenges, we formulate a multi-role RL system where three (groups of) policy networks act as specialized agents: a task routing agent incorporates an attention-based similarity recognition module to determine source-target transfer pairs via attention scores; a knowledge control agent determines the proportion of elite solutions to transfer; and a group of strategy adaptation agents control transfer strength by dynamically controlling hyper-parameters in the underlying EMT framework. Through pre-training all network modules end-to-end over an augmented multitask problem distribution, a generalizable meta-policy is obtained. Comprehensive validation experiments show state-of-the-art performance of our method against representative baselines. Further in-depth analysis not only reveals the rationale behind our proposal but also provide insightful interpretations on what the system have learned.     
\end{abstract}


\section{Introduction}
\IEEEPARstart{E}{volutionary} Algorithms~(EAs) are meta-heuristics that resemble Darwinian principle~\cite{darwinian}, where solution population goes through reproduction and natural selection to search for optima~in given optimization problems. Traditional EAs, such as Genetic Algorithm~(GA)~\cite{GA}, Differential Evolution~(DE)~\cite{DE} and their modern variants~\cite{EA-Survey1,EA-Survey2}, have long-standing reputation for addressing optimization problems in both industrial and scientific discovering fields~\cite{2012opt-economics-survey,2016opt-engi-survey,2018opt-discovery}. Recently, a novel scenario termed as Multitask Optimization~(MTO) was introduced by Gupta et al~\cite{refs:MFEA} to address the urgent need for efficient algorithms that can solve concurrent tasks in cloud computing industry. The complex nature of MTO poses new challenges to traditional EAs: how to realize effective knowledge transfer mechanism to accelerate the optimization process when handling multiple tasks simultaneously~\cite{mto-challenges,refs:MTO-DRA}? 

To address the above challenges, Evolutionary Multitasking~(EMT) has emerged as a prominent solution, with its approaches generally falling into two branches according to their underlying paradigms. The first category is EMT algorithms with implicit knowledge transfer mechanism, first proposed by Gupta et al.~\cite{refs:MFEA} along with the proposal of MTO definition. The proposed algorithm, termed as Multifactorial Evolution Algorithm~(MFEA), maintains a single solution population for the sub-tasks in MTO, where each individual is indexed by its most specialized task. The knowledge transfer occurs within the reproduction~(mutation \& crossover) and selection of the evolution process. While such implicit parallelism facilitates efficient knowledge sharing and concurrent multitasking, it, to some extent, sacrifices the algorithmic flexibility in terms of customization for each sub-task~\cite{refs:EBS}. This motivated the second category of researches that explore EMT algorithms with explicit knowledge transfer mechanisms. The paradigm in this line deploys a group of customized evolution processes for the sub-tasks in MTO. The knowledge transfer is governed by an explicit measure of inter-task similarity and an associated information exchange operator across different evolution processes. For example, in an early-stage study of Feng et al.~\cite{refs:EMT_autoencoding}, denoising autoencoders are trained for inter-task solution mapping to dynamically transfer solution-level knowledge. In this paper, we focus on the second paradigm of developing explicit EMT algorithms.

The development of explicit EMT algorithms centers on three fundamental questions regarding the knowledge transfer: 
\begin{itemize}
    \item \textbf{Where to Transfer:} Identifying which tasks should exchange information, typically by measuring inter-task similarity~\cite{refs:MaTEA}.
    \item \textbf{What to Transfer:} Determining the specific knowledge (e.g., proportion of solutions) to be conveyed from a source task to a target task~\cite{refs:MTO-DRA}.
    \item \textbf{How to Transfer:} Designing the precise mechanism for knowledge exchange between correlated tasks, such as the selection of evolutionary operators~\cite{refs:explicit-AOS} or the control of transfer intensity~\cite{refs:explicit-DAC}.
\end{itemize}
Following these research questions, a wide array of explicit EMT algorithms have been proposed~\cite{refs:AEMTO,refs:EMaTO-MKT}. 

Though promising, the performance of existing EMT algorithms might be cursed by \emph{no-free-lunch} theorem~\cite{NFL}. This theorem implies that an algorithm with fixed, manually-designed components, while excelling in some optimization environments, will inevitably struggle when faced with novel problems or unexpected shifts in the problem landscape. Consequently, adapting or redesigning such algorithms demands significant expertise and development effort, a challenge frequently highlighted in recent surveys on Meta-Black-Box Optimization~(MetaBBO)~\cite{metabbo-survey-1, metabbo-survey-2,metabox, metabox-v2}. The MetaBBO paradigm proposes automating the algorithm design workflow through meta learning~\cite{meta-survey,2019metalearning}. This is typically achieved by a bi-level architecture, where a meta-level policy is trained to configure or generate a low-level algorithm, in order to boost the overall performance of the low-level optimization over a problem distribution, thereby generalizable to unseen problems. While the MetaBBO paradigm has been widely instantiated in several key avenues of EAs community such as single-objective optimization~\cite{configx,pom,metade}, multi-objective optimization~\cite{xue2022madac,rlmoo}, and multi-modal optimization~\cite{lian2024rlemmo,rlmmo}, to the best of our knowledge, only two works~\cite{refs:RLMFEA,refs:L2T} explore the possibility of learning-boost EMT algorithms. While these initial works use reinforcement learning to address knowledge transfer, they fall short of providing holistic control, automating only isolated aspects of the ``where, what and how'' decisions. The lack of comprehensive and coordinated strategy leads to two critical issues: it not only results in suboptimal designs and performance bottlenecks, but it also fails to fully resolve the reliance on human expertise that MetaBBO aims to address. 

\begin{figure}[t]
\centering
\includegraphics[width=0.95\columnwidth]{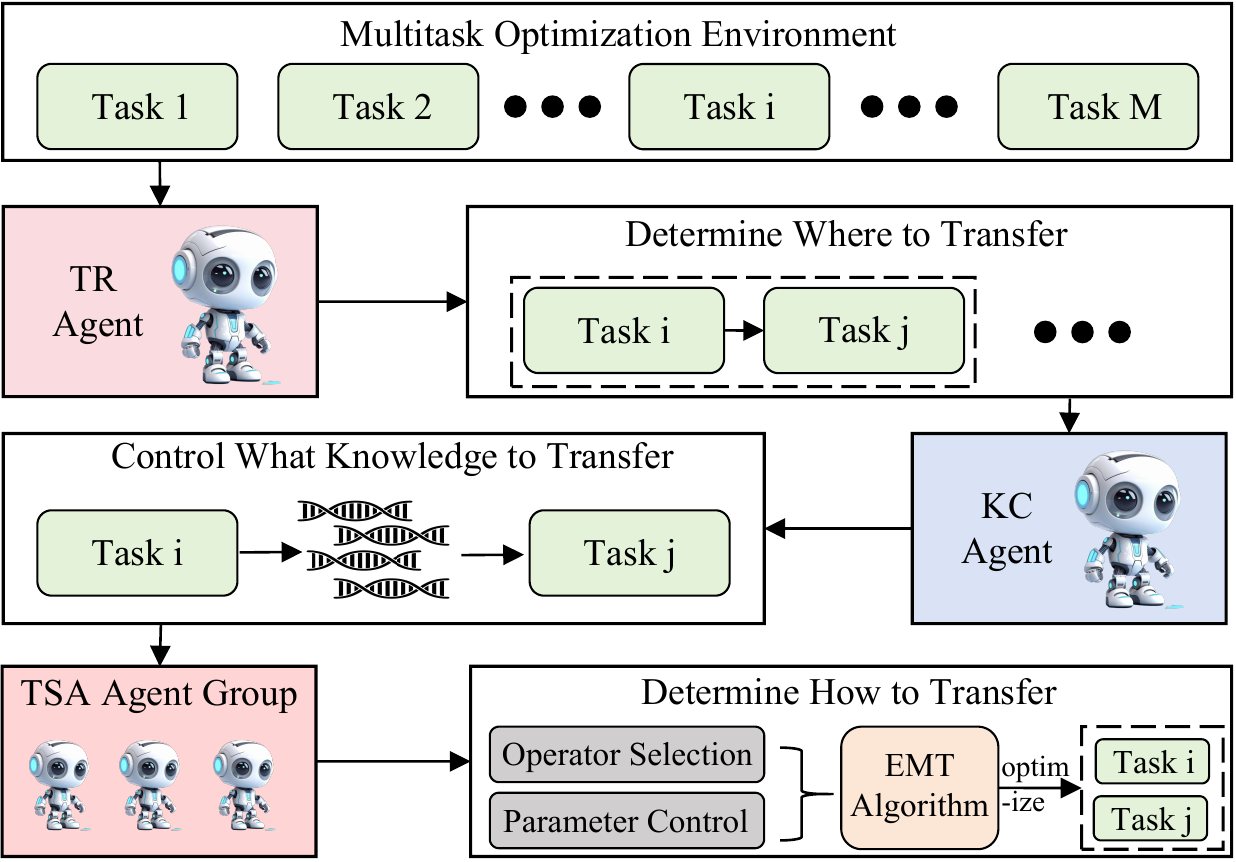}
\caption{Conceptual overview of \textbf{\texttt{MetaMTO}} framework. }
\label{fig:intro}
\end{figure}
To fill the research gap mentioned above, in this paper, we propose \textbf{\texttt{MetaMTO}}, a novel MetaBBO framework designed to learn systematic and generalizable EMT algorithms. Our approach is centered directly on the three fundamental questions of EMT: determining the source task (\underline{\textbf{where}}), the knowledge to be transferred (\underline{\textbf{what}}), and the method to transfer (\underline{\textbf{how}}). To this end, in \textbf{\texttt{MetaMTO}}, we propose an end-to-end reinforcement learning~(RL)~\cite{rl-sutton} system to learn a cohesive policy that concurrently addresses all three interconnected questions. We illustrate our work's novelty in Fig.~\ref{fig:intro}, where RL agents with various roles are holistically organized together to facilitate fully automated control for knowledge transfer. Specifically, three types of agents are deployed to govern knowledge transfer in a cooperative fashion:
\begin{itemize}
    \item \textbf{Task Routing (TR) Agent}: This agent addresses the ``where to transfer" question. It processes status features extracted from all sub-tasks and uses an attention-based architecture to compute pairwise similarity scores. Based on the attention scores, it routes each target task with the most-matching source task for knowledge transfer.
    \item \textbf{Knowledge Control~(KC) Agent}: : This agent is responsible for the ``what to transfer" decision. For each recognized source-target pair identified by the TR agent, the KC agent determines the quantity of knowledge to transfer by selecting a specific proportion of elite solutions from the source task's population.
    \item \textbf{Transfer Strategy Adaption~(TSA) Agent Group}: These agents tackle the final ``how to transfer" question. For each source-target pair and corresponding knowledge to be transferred, a TSA agent further suggest the desired transfer strategy, through controlling the key algorithm configurations in the underlying EMT framework.
\end{itemize}

At its core, \textbf{\texttt{MetaMTO}} frames the EMT algorithm control as a Markov Decision Process (MDP) and employs a multi-role RL system to solve it, which enables a dynamic, per-step policy to control the entire knowledge transfer process for each MTO problem. To facilitate the training and evaluation, we further make following contributions to complement the overall systematic pipeline, including i) \textit{Training Objective}: a novel reward scheme is designed to objectively measure how good \textbf{\texttt{MetaMTO}} is at designing per-step knowledge transfer strategy, kicking a balance between global convergence performance and transfer success rate. Then the training objective is maximizing the accumulated rewards along an optimization process; ii) \textit{Training Diversity}: a novel MTO problem set is constructed by augmenting existing ones~\cite{refs:CEC2017_WCCI2020}, which allows diversified problem distribution through efficient hierarchical composition. Meta-training \textbf{\texttt{MetaMTO}} over such training data leads to competitive generalization. 

We have conducted rigorous comparative validation, where the results demonstrate state-of-the-art performance of our work against both human-crafted and learning-assisted baselines. Plentiful ablation studies and interpretation analysis further enhance our work's scientific integrity. 

The remaining of this paper is organized as follows. In Section~\ref{sec:2}, we systematically review existing human-crafted EMT algorithms and learning-assisted ones, highlighting the research gap in existing research progress. Section~\ref{sec:3} present a step-by-step derivation of \textbf{\texttt{MetaMTO}}, including both the mathematical modeling and specific technical designs. Thereafter, in Section~\ref{sec:4}, rigorous validation studies are conducted to empirically support our work's significance, which include not only absolute performance comparison but also in-depth ablation and interpretability analysis. To conclude, the main contributions and findings of the paper are summarized in Section~\ref{sec:6}, along with several promising directions we outline as important future works.

\section{Related Work}\label{sec:2}
\subsection{Evolutionary Multitasking}\label{sec:2.1}
Evolutionary multitask optimization~(EMT) is specifically designed to solve multitask optimization~(MTO) problems~\cite{refs:MFEA}. It aims to  solve multiple sub-tasks synergistically within a single optimization process. The objective for a given MTO problem instance $\mathcal{I}:=\left\{T_1, T_2, \ldots, T_K\right\}$ can be mathematically defined as finding the set of optimal solutions $\left\{x_1^*, x_2^*, \ldots, x_K^*\right\}$, where each solution is given by:

\begin{equation}\label{eq:mto}
    x_j^* = \underset{x \in R_j}{\mathrm{argmin}}f_j(x), \quad j = 1,2,...,K,
\end{equation}
where $f_j$ and $R_j$ denote the objective function and feasible region of sub-task $T_j$. MTO presents significant challenges to traditional EAs due to i) heterogeneous landscape properties of the objective functions $\{f_j\}_{j=1}^K$ across sub-tasks; and ii) misaligned feasible decision variable regions $\{R_j\}_{j=1}^K$. Obviously, the similarity and dissimilarity between sub-tasks play a key clue in MTO and should be addressed carefully. Motivated accordingly, existing EMT approaches mainly focus on designing effective \emph{knowledge transfer} scheme. The fundamental rationale is that by distinguishing similar and dissimilar sub-tasks, we could assign computational resources properly to attain the optimality of MTO more efficiently. Following this idea, a wide range of EMT approaches have been proposed recently and they can be categorized into two branches according to their specific knowledge transfer paradigms~\cite{refs:EMT_survey}.

\subsubsection{Implicit Knowledge Transfer}
In the implicit knowledge transfer scheme, a unified population is typically maintained to address all sub-tasks. Knowledge transfer is then implicitly conducted within the reproduction and selection of the evolution process. In the realm of implicit EMT algorithms, the pioneering work of Multifactorial Evolutionary Algorithm (MFEA) was first introduced by Gupta et al. \cite{refs:MFEA}. In MFEA, each individual is assigned a skill factor to indicate its adept task. The occurrence of implicit knowledge transfer is controlled by a fixed random mating probability~($rmp$) between individuals with distinct skill factors. Recognizing that the fixed $rmp$ in MFEA has limited capability in dynamically adjusting knowledge transfer probability and may lead to negative knowledge transfer, Bali et al.~\cite{refs:MFEA-II} introduced a data-driven evolutionary multitasking algorithm (MFEA-II). MFEA-II employs an adaptive $rmp$ matrix to represent pairwise transfer probabilities between tasks, which can be updated online using real-time data. Furthermore, to more effectively control the occurrence of knowledge transfer, Zheng et al.~\cite{refs:SREMTO} designed an ability vector for each individual to indicate its probability of being selected as transferred knowledge. Tang et al.~\cite{refs:GMFEA} proposed a grouping strategy that clusters similar tasks together, permitting inter-task knowledge transfer only within the same group. 

The aforementioned works consider ``where'' and ``what'' to transfer in implicit EMT, whereas some other efforts pay attention to ``how'' to transfer the knowledge. Bali et al.~\cite{refs:LDA-MFEA} proposed a linear domain adaptation strategy to transform the search space of a simple source task to the complex target task. Furthermore, Zhou et al.~\cite{refs:MFEA-AKT} proposed an adaptive knowledge transfer approach, which assigns an operator indicator for each individual to adaptively select the most suitable crossover operators. Li et al.~\cite{refs:MFEA-TLS} incorporated a team learning strategy into the MFEA. They divides the population into one elite and three ordinary teams, with each team conducting knowledge transfer by a distinct approach.  

\subsubsection{Explicit Knowledge Transfer}
While implicit EMT algorithms have shown promising results in diverse optimization domains~\cite{refs:app_multiobjective,refs:app_large_scale}, they still face some limitations, such as the insufficient flexibility for per-task adaptation and the limited control over dynamic resource allocation. The explicit knowledge transfer was then proposed by Feng et al.~\cite{refs:EMT_autoencoding} and has since become an active area of research. Specifically, the method is characterized by its multi-population structure, where each sub-task is optimized by a dedicated population. This allows more flexible adaptation within the separated sub-task domain, while also enabling explicit transfer operations between them through well-defined operations like solution exchange and transformation. To explore the where to transfer question, population distribution-based similarity measurement approach is often used. Chen et al.~\cite{refs:MaTEA} proposed to employ an archive for each task to record previous and current population and select the auxiliary task by measuring the Kullback-Leibler divergence~\cite{refs:KLD} for task similarity between paired task archives. Along the line, Wang et al.~\cite{refs:MTEA-AD} introduced an anomaly detection model fitted by the population of each task to detect candidate transferred individuals.  Gong et al. \cite{refs:MTO-DRA} proposed MTO-DRA, which incorporates dynamic resource allocation into explicit knowledge transfer. This approach dynamically assigns more computational resources to more challenging tasks, thereby regulating the amount of transferred knowledge. 

Finally, the question of how to transfer has been approached. Lin et al.~\cite{refs:explicit-AOS} focused primarily on the selection of information exchange operators. They applied an indicator vector to dynamically choose domain adaptation approaches for learned mappings between different task domains. In parallel, Wu et al.~\cite{refs:explicit-DAC} studied the transfer strength control and introduced an approach termed TRADE. This approach transfers valid DE parameter settings from auxiliary to target tasks as explicit knowledge, thereby enhancing the search efficiency. From another perspective, Jiang et al.~\cite{refs:BLKT} proposed a block-based mechanism and enabled explicit knowledge transfer to occur in non-aligned dimensions of individuals of the same or different tasks. In addition, some recent researches develop integrated approaches. Xu et al.~\cite{refs:AEMTO} proposed an adaptive EMT framework that coordinates knowledge transfer frequency, auxiliary task selection and transfer intensity in a synergistic manner. In a similar vein, Liang et al. \cite{refs:EMaTO-MKT} introduced a multi-source knowledge transfer mechanism, which integrates  novel designs for adjusting transfer probability, identifying auxiliary tasks, and facilitating the knowledge transfer process. 

The EMT approaches reviewed above have shown promising results in real-world MTO scenarios~\cite{refs:app_real_world}, e.g., the vehicle routing problem~\cite{refs:app_vrp1,refs:app_vrp2}, the  itinerary planning~\cite{refs:S2E-GA} and the fuzzy system design~\cite{refs:app_fuzzy_system}.

\subsection{Learning-assisted Evolutionary Multitasking}\label{sec:2.2}

Though promising, a key limitation of human-crafted EMT approaches is that: being designed and tuned for a specific group of problems, they unavoidably require labor-intensive fine-tuning or even re-design to adapt for new problems. This issue has been well summarized as the ``\emph{no-free-lunch}'' curse in recent learning-assisted optimization literature, where MetaBBO are suggested as effective tools to approach generalized optimization~\cite{metabbo-survey-1,metabbo-survey-2,neurela,configx,designx,surr-rlde,qmamba,rldealf,lsre}. Motivated by this perspective, Chen et al.~\cite{refs:SL_ANN} and Xue et al.~\cite{refs:SL_NNMTO} both employed neural networks to model transfer mappings between task pairs. While Chen et al. used zero padding to align dimensions across tasks, Xue et al. directly mapped solutions even between tasks with different dimensions. On the other hand, to mitigate the computational burden caused by additional function evaluations in multitask optimization, surrogate-assisted EMT algorithms have been developed. Huang et al.~\cite{refs:SL_SATS-EMO} proposed a novel surrogate-assisted task selection approach. In this approach, a surrogate model is trained to determine the skill factor of each offspring and identify the most promising candidates for real and precise evaluation. 

The above studies represent significant progress at the component level, by designing sophisticated transfer mechanisms and efficient heuristics. However, a higher-level approach is required to shift the focus from designing the transfer mechanisms to learning the meta-level policy that governs them. This aligns with the essential goal of the MetaBBO paradigm, which, to our knowledge, has received little attention so far. Li et al. \cite{refs:RLMFEA} introduced an adaptive strategy using the Q-learning method to dynamically optimize the $rmp$ value for each task. Wu et al. \cite{refs:L2T} developed a learning-to-transfer~(L2T) framework that autonomously regulates the knowledge transfer process and effectively addresses the issues of when and how to transfer across tasks. 

Despite their promise, existing learning-assisted EMT works are still limited at the early exploratory stage. Taking L2T~\cite{refs:L2T} as an example, on the one hand, its rigid architecture is tied to a fixed number of tasks, which hinders generalization. On the other hand, the provided control is in-holistic, leaving the ``where'' to transfer mechanism unexplored. This motivates us to propose \textbf{\texttt{MetaMTO}} for more systematic and generalized learning to holistically address ``where'', ``what'' and ``how'' questions in an end-to-end manner.

\section{Methodology}\label{sec:3}
\begin{figure*}[ht]
\centering
\includegraphics[width=0.99\linewidth]{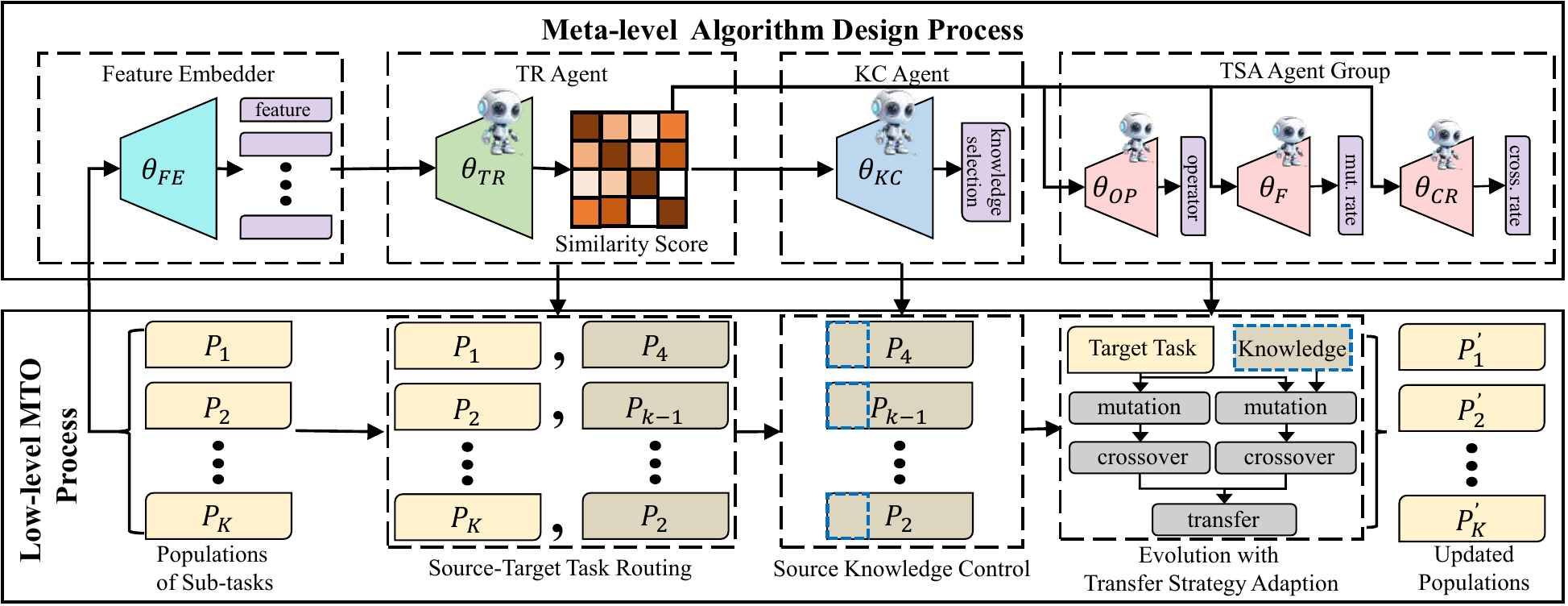}
\caption{A detailed illustration of the overall workflow within \textbf{\texttt{MetaMTO}}. Following the bi-level architecture in MetaBBO, a multi-role RL system is deployed at the meta-level to control the knowledge transfer scheme at the low-level evolution. A single evolution step is presented for simplicity.}
\label{fig:method}
\end{figure*}

As illustrated in Fig.~\ref{fig:method}, \textbf{\texttt{MetaMTO}} aligns with the bi-level MetaBBO paradigm~\cite{metabbo-survey-1}, while equipped with novel designs tailored for EMT backbone to solve MTO problems. We begin with modeling the interplay between the meta-level policy and the low-level EMT process and defining the corresponding learning objective. Without loss of generality, we assume that all sub-tasks in an MTO problem hold minimization objectives, and the underlying EMT approach could be any off-the-shelf multi-population EAs, which in this paper, is a primitive multi-population DE architecture that resembles initial EMT researches~\cite{refs:EBS,refs:MaTEA} with explicit knowledge transfer schemes. 

\subsection{Problem Modeling}
Suppose we have a MTO problem distribution $\mathbb{D}$ from which diverse MTO problem instances can be sampled. Each MTO instance $\mathcal{I}$ sampled from $\mathbb{D}$ comprises $K$ sub-tasks $T_1 \sim T_K$. Our aim is to learn an algorithm design policy $\pi_\theta$, where $\theta$ are some learnable parameters, to control the evolutionary optimization process of a given EMT approach $\mathcal{A}$, so that it achieves desired optimization performance on each $\mathcal{I} \in \mathbb{D}$. We now clarify the specific mathematics behind via diving into the low-level evolutionary process. During the optimization of each $\mathcal{I}$ by the EMT approach $\mathcal{A}$, we first define $\mathcal{R}_\mathcal{I}^t$ as a scalar performance metric capable of profiling the optimization progress information as of $t$-th optimization step  across all $K$ sub-tasks in $\mathcal{I}$~(larger is better). In existing MetaBBO approaches tailored for single-task optimization, the design of $\mathcal{R}_\mathcal{I}^t$ majorly focuses on objective's optimality. However, in this work, since we aim to address MTO problems, the design has to consider optimality in each sub-task's objective space and also the knowledge transfer status, which we detail in Sec~\ref{sec:3.4}. Following this design principle, at each time step $t$, a state function $\text{sf}(\cdot)$ maps the meta information~(e.g., populations and objective values for sub-tasks, knowledge transfer history, etc.) into a state representation $s^t$. The policy $\pi_\theta$ then conditions on state $s^t$ to output a knowledge transfer action $a^t$ for $\mathcal{A}$. By using $a^t$ to optimize $\mathcal{I}$, we advance the evolutionary process by one step and record the performance metric $\mathcal{R}_\mathcal{I}^{t}$. The learning objective of \textbf{\texttt{MetaMTO}} is to find a policy that maximizes the expectation of accumulated $\mathcal{R}_\mathcal{I}^{t}$ over the MTO problem distribution $\mathbb{D}$:
\begin{equation}\label{eq:mdp}
    \mathcal{J}(\theta) = \underset{\mathcal{I} \in \mathbb{D}}{\mathbb{E}} \left[ \sum_{t=1}^{G} \mathcal{R}_\mathcal{I}^{t} \right]
\end{equation}
where $G$ denotes the number of allowed maximal optimization steps for each problem instance. With this problem modeling, we next detail designs in \textbf{\texttt{MetaMTO}} for each mathematical component above.  

\subsection{MDP Formulation}

We formally model the above meta-learning problem as a Markov Decision Process~(MDP)~\cite{markov}, where the formalized $s^t$ denotes state, the configuration decision $a^t$ denotes action, and the immediate performance metric $\mathcal{R}_\mathcal{I}^{t}$ denotes the reward. We detail their concrete designs as follows.
\begin{table}[t]
	\caption{State definition in \textbf{\texttt{MetaMTO}} for each sub-task population.}	
    \centering
    \resizebox{0.9\columnwidth}{!}{
	\begin{tabular}{c l c l }
		\toprule[2pt]
		  State ID & Definition & Type & Description\\
		\midrule 
		$s_1$ & $\text{mean}(\text{std}(X))$ & $[0,1]$ & diversity in decision space\\
        \specialrule{0em}{1pt}{1pt}
        $s_2$ & $\text{std}(f(X))$ & $[0,1]$ & convergence in objective space\\
        \specialrule{0em}{1pt}{1pt}
        $s_3$ & $\frac{G_{stg}}{G}$ & $[0,1]$ & stagnation status in optimization\\
        \specialrule{0em}{1pt}{1pt}
        $s_4$ & $\mathrm{I}(f(X^{t+1})<f(X^t) )$ & $bool$ & new best-so-far solution checking\\
        \specialrule{0em}{1pt}{1pt}
        $s_5$ & $\frac{N_{sur}}{N_{tra}}$ & $[0,1]$ & survival rate of transferred solutions\\

		\bottomrule [2pt]
	\end{tabular}
    }
    \label{tab:state}
\end{table}

\begin{table}[t]
	\caption{Action groups in \textbf{\texttt{MetaMTO}} for effective knowledge transfer.}	
    \centering
    \resizebox{0.99\columnwidth}{!}{
	\begin{tabular}{l c c l }
		\toprule[2pt]
		  Group & Action ID & Range & Description\\
		\midrule 
		\textsc{task routing} & $\{a_{(1),j}\}_{j=1}^{K}$ & $\{1,2...,K\}^K$ & assign source task for the recipient task\\
        \midrule 
        \specialrule{0em}{1pt}{1pt}
        \textsc{Knowledge Selection} & $\{a_{(2),j}\}_{j=1}^{K}$ & $[0,0.5]^K$ & determine the elite portion to transfer\\
        \midrule 
        \specialrule{0em}{1pt}{1pt}
        \multirow{3}{*}{\textsc{Transfer Strategy Adaption}} & $\{a_{(3.1),j}\}_{j=1}^{K}$ & $\{1,2,3,4\}^K$ & dynamic mutation operator selection\\
        \specialrule{0em}{1pt}{1pt}
         & $\{a_{(3.2),j}\}_{j=1}^{K}$ & $[0,1]^K$ & control mutation strength\\
        \specialrule{0em}{1pt}{1pt}
         & $\{a_{(3.3),j}\}_{j=1}^{K}$ & $[0,1]^K$ & control crossover strength\\

		\bottomrule [2pt]
	\end{tabular}
    }
    \label{tab:action}
\end{table}

\subsubsection{State}\label{sec:state}
At each optimization step $t$ of the underlying EMT process, given the populations $\{(X_j^t,f_j(X_j^t))\}_{j=1}^{K}$ for each of the $K$ sub-tasks, we construct for each sub-task a 5-dimensional feature vector $s_j^t$ to describe the optimization state information. We present the mathematical definition of these feature logits in Table~\ref{tab:state}, where we omit subscripts $j$ and superscript $t$ wherever appropriate. Note that such state feature design not only coheres with the up-to-date MetaBBO researches~\cite{surr-rlde,rldealf,pom-v2,designx,refs:L2T} but also further incorporates the optimization information of knowledge transfer. Specifically, according to recent landscape feature analysis literature~\cite{ela-1, ela-2, neurela, deepela}, we first distill a concise set of features to avoid redundancy. This results in four representative features~($s_1$ to $s_4$) that cover the diversity, convergence, stagnation and update process. Then, inspired by L2T~\cite{refs:L2T}, we purposely add a key state feature~($s_5$) that quantifies the positive impact of the knowledge transferred to the recipient sub-task's evolution. The incorporation of $s_5$ enables the agent to characterize both task-specific and cross-task optimization dynamics, thereby facilitating comprehensive and coordinated action-making for knowledge transfer. The complete state feature is hence the union of state features $s^t := \{s_j^t\}_{j=1}^{K}$ from all $K$ sub-tasks. We leave the computation details of these features at Appendix I.A. 

\subsubsection{Action}\label{sec:action}

To provide a comprehensive and sophisticated control, an agent must answer three fundamental questions of where, what, and how to transfer. Listed in Table~\ref{tab:action},
we structure our action space  by defining three distinct action groups, each responsible for one of these decisions. The action space is formulated as $\textbf{A}:=\{a_{(1)},a_{(2)},a_{(3.1)},a_{(3.2)},a_{(3.3)}\}$, where action groups $a_{(1)}$, $a_{(2)}$ and $a_{(3)}:=\{a_{(3.1)},a_{(3.2)},a_{(3.3)}\}$ respectively correspond to task routing, knowledge control and transfer strategy adaption: 
\begin{itemize}
    \item \textbf{Task Routing (TR)}: In this group, each of the $K$ sub-tasks is regarded as a target task and to be assigned with a source task. Then the overall action is $a_{(1)} := \{a_{(1),j}\}_{j=1}^{K}$, with each $a_{(1),j}$ is an integer index selected from $\{1,2, \ldots, K\} \backslash\{j\}$~(for $j$-th sub-task, index $j$ is not allowed to be selected). This action group is responsible for determining where to transfer knowledge.
    \item \textbf{Knowledge Control (KC)}: Once the source task is assigned, we then determine what knowledge to transfer. This is achieved by  determining a proportion $a_{(2),j}\in[0,0.5]$, and then selecting that proportion of elite solutions from source task's population. The overall action is hence $a_{(2)} := \{a_{(2),j}\}_{j=1}^{K}$. Note that the upperbound is set to 0.5 to prevent excessive knowledge transfer that may pollute the original evolution of the target task. 
    \item \textbf{Transfer Strategy Adaption (TSA)}: We further propose three action sub-groups $a_{(3.1)}$, $a_{(3.2)}$ and $a_{(3.3)}$ to provide fine-grained algorithmic control over the low-level evolution algorithm (which is a multi-population DE architecture in this study). Specifically, we control the mutation operator selection, mutation strength $F$, and the crossover probability $Cr$, respectively, to provide per-step control for the DE. For each $a_{(3.1),j}$, it is selected from a predefined operator pool comprising four behavior-varied mutation strategies. As listed in Table~\ref{tab:operator}, these four operators provide different extent of exploration-exploitation tradeoff between the target and source tasks, which are dynamically selected within the EMT process to facilitate adaptive optimization. The other two sub-groups control the parameters of the mutation and crossover operators\footnote{In this paper, we do not control the selection of crossover operator, using binomial crossover by default.}. Specifically, the action $a_{(3.2),j}$ determines the mutation strength $F \in [0,1]$, and the action $a_{(3.3),j}$ sets the crossover probability $Cr \in [0,1]$, respectively, for the $j$-th sub-task. 
\end{itemize}
 \begin{table}[t]
	\caption{Four different operators in the mutation operator pool.}	
    \centering
    \resizebox{0.85\columnwidth}{!}{
	\begin{tabular}{c c }
		\toprule[2pt]
		   Operator ID & Formulation \\
		\midrule 
		1 & $\mathcal{V} = X_{\text{target},\text{best}} + F\cdot(X_{\text{source},r_1}-X_{\text{source},r_2})$ \\
        \specialrule{0em}{1pt}{1pt}
        2 & $\mathcal{V} = X_{\text{target},r_1} + F\cdot(X_{\text{source},r_2}-X_{\text{source},r_3})$ \\
        \specialrule{0em}{1pt}{1pt}
        3 & $\mathcal{V} = X_{\text{source},r_1} + F\cdot(X_{\text{target},r_2}-X_{\text{target},r_3})$ \\
        \specialrule{0em}{1pt}{1pt}
        4 & $\mathcal{V} = X_{\text{source},best} + F\cdot(X_{\text{target},r_1}-X_{\text{target},r_2})$ \\
		\bottomrule [2pt]
	\end{tabular}
    }
    \label{tab:operator}
\end{table}

In summary, there are three action groups in \textbf{\texttt{MetaMTO}}, which cooperate to achieve systematic and fine-grained per-step control for the low-level EMT framework. 

\subsubsection{State Transition Dynamics}\label{sec:dynamics}
Given the state and action definitions, the state transition dynamics in our MDP model correspond to one iterative evolution step of the underlying EMT framework~(as illustrated in the bottom right of Fig.~\ref{fig:method}). Specifically, once the meta-level policy provides the action $a^t$ conditioning on the current state  $s^t$, the underlying EMT framework performs a mutation-crossover routine to generate an offspring population of $\{V_j^t\}_{j=1}^{K}$. First, a proportion of the offspring, corresponding to $(1-a_{(2), j}^t)$ of the population size, is generated through self-evolution in the $K$ sub-tasks by utilizing the local information within population. Then, the remaining $a_{(2), j}^t$ proportion of population is generated by cross-task knowledge transfer. Guided by the other components of the action $a^t$, the pathway involves: identify the source task by $a_{(1), j}^t$, and adaptively transfer knowledge between the selected elites and local population with behaviors controlled by $a_{(3.1)}^t$, $a_{(3.2)}^t$ and $a_{(3.3)}^t$. Finally, to produce the next generation, each sub-task's parent population and offsprings undergo a vanilla DE selection strategy, where the fitter individual from each parent-offspring pair survives to the next generation. The resulting population provides the next state information, $s^{t+1}$, in our MDP. 

\subsubsection{Reward}\label{sec:3.4}
Reward design plays an important role in RL systems, which impacts the overall training robustness and effectiveness~\cite{reward-design}. It is also reported in MetaBBO literature that designing robust reward signal for learning-assisted optimization is more challenging than typical RL tasks~\cite{metabbo-survey-1}. To strike a balance between each sub-task's self optimization and the exploitation of knowledge transfer, we design a reward function by crediting two terms. This design not only promotes maximal performance gain in each sub-task's evolution, but also encourages positive knowledge transfer. The reward function is defined as:

\begin{equation}\label{eq:reward}
\begin{aligned}
    &\mathcal{R}^t = \sum_{j=1}^{K} (\mathcal{R}_{c,j}^t + \mathcal{R}_{k,j}^t), \\
    &\mathcal{R}_{c,j}^t = \frac{f_j^t - f_j^{t+1}}{f_j^0 - f_j^*}, \mathcal{R}_{k,j}^t = \frac{N_{\text{success},j}^t}{N_{\text{transfer},j}^t}.
\end{aligned}
\end{equation}
where $f_j^*$ denotes the optimal of $j$-th sub-task, $f_j^t$ denotes the optimal found until the $t$-th optimization step, $N_{\text{transfer},j}^t$ denotes the number of all transferred solutions from source task and $N_{\text{success},j}^t$ denotes the survived number.  It should be noted that for black-box MTO, the true $f_j^*$ is agnostic. In such case, $f_j^*$ can be replaced by the proxy optimal, which is the best solution found using existing EMT approaches through multiple runs. This proxy value is exclusively a training-phase construct; it is not required during the inference phase, when the trained controller is applied to solve problems.  

\subsection{Multi-Role RL System}\label{sec:agent}
Previously, we have elaborated the MDP details in our \textbf{\texttt{MetaMTO}}, including the state, action and reward  designs. To solve this MDP, we introduce the meta-level policy: a multi-role RL system comprising multiple RL agents. As illustrated in the top of Fig.~\ref{fig:method}, these agents are instantiated as neural networks. In this subsection, we elaborate the data flow in the system, explaining how these agents operate collectively.

\subsubsection{Feature Embedder}
Given the abstracted state feature $s^t := \{s_j^t\}_{j=1}^{K}$, where $s_j^t \in \mathbb{R}^5$ is the state feature for $j$-th sub-task at optimization step $t$. We first use a feature embedder $\mathcal{W}(\cdot|\theta_{\text{FE}})$ to embed them into a hidden feature space through the linear mapping $\theta_{\text{FE}}: \mathbb{R}^5 \rightarrow \mathbb{R}^{64}$. Then we obtain embeddings $e^t = \mathcal{W}(s^t|\theta_{\text{FE}})$ for all sub-tasks, where $e^t \in \mathbb{R}^{K \times 64}$. The embeddings are used by multi-role RL agents to determine the corresponding action groups. 

\subsubsection{TR Agent}
To determine where to transfer the knowledge, the feature embeddings are first fed into the TR agent, which is an attention block~\cite{transformer}, represented by $\mathcal{W}(\cdot|\theta_{\text{TR}})$. The block contains a single attention head with $64$-dimentional hidden state, followed by a batch normalization~\cite{BN}. Given $e^t$, the TR agent outputs two logits:
\begin{equation}\label{eq:embed}
    h_{\text{score}}^t, h_{\text{decision}}^t = \mathcal{W}(e^t|\theta_{\text{TR}})
\end{equation}
where $h_{\text{score}}^t \in \mathbb{R}^{K \times K}$ denotes the attention score matrix to be used immediately in current task routing stage, and $h_{\text{decision}}^t \in \mathbb{R}^{K \times 64}$ is a feature matrix to be passed to subsequent agents. 

First, the attention scores are used to pair each target task $j$ with a source task $a_{(1), j}^t$ by selecting the task with the highest score (excluding the task itself):

\begin{equation}
    a_{(1),j}^t = \underset{k \in \{1,2...,K\}}{\mathrm{argmax}} h_{\text{score}}[j,k], \quad k \neq j. 
\end{equation}
The resulting target-source pairs $\{j,a_{(1),j}^t\}_{j=1}^{K}$ are used in the knowledge transfer scheme of the low-level EMT framework. 

Next, using these routing pairs, we construct an enriched input for subsequent RL agents. This is achieved by concatenating the decision embedding of each target task $j$ with that of its assigned source task $a_{(1), j}^t$:
\begin{equation}
    h_{\text{concat,j}}^t = \text{Concat}(h_{\text{decision}}^t[j],h_{\text{decision}}^t[a_{(1),j}^t])
\end{equation}
This concatenation yields the final input $h_{\text{concat}}^t \in \mathbb{R}^{K \times 128}$ to the next stage.

\subsubsection{KC Agent}
The KC agent determines what knowledge~(how much percentage of elite solutions) should be transferred from the source task $a_{(1),j}^t$ to the target $j$. Its architecture is a two-layer MLP with hidden layer's dimension as $64$~(activated by ReLU) and the output layer's dimension as $1$~(activated by Tanh). The agent is represented as $\mathcal{W}(\cdot|\theta_{\text{KC}})$. The primary design goal is to generate a transfer ratio, $a_{(2), j}^t$, bounded within the range $[0,0.5]$ to prevent ``knowledge pollution''. Specifically, fed with $h_{\text{concat,j}}^t$, the KC agent outputs a mean value:
\begin{equation}\label{eq:kc}
    \mu_{\text{kc},j}^t = 0.25 + 0.25\times \mathcal{W}(h_{\text{concat,j}}^t|\theta_{\text{KC}})
\end{equation}
Since the MLP's final Tanh activation ensures $\mathcal{W}\left(\cdot \mid \theta_{\mathrm{KC}}\right) \in [-1,1]$, the $\mu_{\mathrm{kc}, j}^t$ is inherently restricted to the desired $[0,0.5]$. Then the concrete percentage of transferred knowledge is sampled from a Gaussian distribution $a_{(2),j}^t \sim \mathcal{N}(\mu_{\text{kc},j}^t, 0.1)$ centered at this bounded mean, where we fix the variance term as $0.1$ for  simplicity. To strictly enforce the boundary constraint, any sampled value falling outside the $[0,0.5]$ interval is truncated to fit within this range.

\subsubsection{TSA Agent Group}
To further determine how to transfer selected knowledge, the first agent in the TSA group is responsible for selecting a proper mutation operator from the operator pool in Table~\ref{tab:operator} for the current step $t$. Its architecture is a two-layer MLP parameterized by $\theta_{\text{OP}}$, with hidden layer's dimension as $64$~(activated by ReLU) and the output layer's dimension as $4$~(activated by ReLU). The operator choice $a_{(3.1),j}^t$ is then sampled by:
\begin{equation}
    a_{(3.1),j}^t \sim \text{Softmax}\left(\mathcal{W}(h_{\text{concat,j}}^t|\theta_{\text{OP}})\right)
\end{equation}
The second and third agents $\theta_{\text{F}}$ and $\theta_{\text{Cr}}$ control the mutation strength $F_j^t$ and $Cr_j^t$ respectively. Their architectures resemble the KC agent, similarly, they also output $1$-dimensional mean terms $\mu_{F,j}^t$ and $\mu_{Cr,j}^t$~(activated by Tanh):
\begin{equation}\label{eq:fcr}
   \mu_{F,j}^t, \mu_{Cr,j}^t = 0.5 + 0.5 \times \mathcal{W}(h_{\text{concat,j}}^t|\theta_{\text{F}},\theta_{\text{Cr}})
\end{equation}
The concrete values are then sampled as $a_{(3.2),j}^t \sim \mathcal{N}(\mu_{F,j}^t,0.1)$ and $a_{(3.3),j}^t \sim \mathcal{N}(\mu_{Cr,j}^t,0.1)$. Likewise, for $a_{(3.2),j}^t$ and $a_{(3.3),j}^t$, we truncate the sampled values from the Gaussian distribution to the bounded range $[0,1]$. 

To summarize, the learnable parameters of \textbf{\texttt{MetaMTO}} are the union of the feature embedder and the five modules' parameters: $\theta := \{\theta_{\text{FE}},\theta_{\text{TR}},\theta_{\text{KC}},\theta_{\text{OP}},\theta_{\text{F}},\theta_{\text{Cr}}\}$. At each optimization step $t$ of the low-level EMT framework, systematic and fine-grained actions are conditionally sampled by the multi-role RL agents for all $K$ sub-tasks' evolution processes: $a^t = \mathcal{W}(s^t|\theta)$. We denote the sample probability is $p_\theta(a^t)$. 

\subsection{Training Method}
\subsubsection{Training Distribution}
Given the MDP and the multi-role RL system already defined, the last piece of our \textbf{\texttt{MetaMTO}} is a well-defined MTO distribution~(problem set), which is crucial to ensure the generalization after the meta-learning~\cite{meta-dataset-1}. Although existing works such as L2T~\cite{refs:L2T} use representative MTO benchmarks such as CEC2017~\cite{CEC2017} or WCCI2020 benchmarks~\cite{WCCI2020}, the benchmarks suffer from two limitations: first, the problem instances in these benchmarks are sparsely and non-uniformly distributed, and second, they contain an insufficient number of instances - often fewer than 10. The data may not be sufficient for a meta-learning agent to learn a truly robust and generalizable policy. 

To address this issue, we develop a more comprehensive problem set, termed as augmented WCCI~(AWCCI), based on the 7 basic functions in WCCI2020~\cite{WCCI2020}. The AWCCI set is constructed in an incremental, hierarchical and automatic fashion. Specifically, we let the instances in AWCCI as the random combinations of those 7 basic functions, which results in $\sum_{i=1}^{7}C_7^i = 127$ possible non-empty combinations. Then, for each combination, we generate 5 distinct MTO problem instances. Each instance consists of 10 parallel, 50-dimensional sub-tasks. The difference of the 5 instances is derived by the shift intensity level $l \in \{0.05, 0.1, 0.2, 0.3, 0.4\}$. Each sub-task in a multitask instance is generated through the following transformation on the basic function defined by its corresponding combination: 
\begin{equation}\label{eq:awcci}
\begin{aligned}
    f^\prime = & f\left(\omega^T \cdot (x-s)\right),\\
    s \sim l\times (\text{lb}+ & U[0,1]\times (\text{ub}-\text{lb}))
\end{aligned}
\end{equation}
where $\omega$ is a random rotation matrix, \text{ub} and \text{lb} denote the searching range of $f$. This process yields a total of $127 \times 5 = 635$ problem instances, which are organized into five datasets according to their shift levels: AWCCI-VS~(with very small shifting), AWCCI-S~(with small shifting), AWCCI-M~(with median shifting), AWCCI-L~(with large shifting) and AWCCI-VL~(with very large shifting). We provide a more detailed elaboration on AWCCI in Appendix I.B.  

\begin{algorithm}[t]
\caption{\textbf{\texttt{MetaMTO}}'s Training Workflow.}
\label{alg:training}
\small
\begin{algorithmic}[1]
\REQUIRE Initialized policy $\pi_{\theta}$, MTO trainset $\mathbb{D_{\text{train}}}$, EMT framework $\mathcal{A}$, Training epochs $L$, Optimization budget $G$.
\STATE set epoch $l = 1$;
\WHILE{$l \leq L$}
    \FOR{each instance $\mathcal{I} \in \mathbb{D_{\text{train}}}$}
        \STATE /* \textcolor{cyan}{\emph{Initialize low-level EMT $\mathcal{A}$}} */
        \STATE  set optimization step $t = 1$;
        \STATE initialize populations $\{X_j^t, f_j^t(X_j^t)\}_{j=1}^{K}$ for $\mathcal{I}$;
        \STATE /* \textcolor{cyan}{\emph{Start optimization}} */
        \WHILE{$t \leq G$}
            \STATE /* \textcolor{cyan}{\emph{RL agent decides on action}} */
            \STATE Get state feature $s^t$~(Sec.~\ref{sec:state});
            \STATE Sample action $a^t \sim \mathcal{W}(s^t|\theta)$~(Eqs.~\ref{eq:embed} $\sim$~\ref{eq:fcr});
            \STATE /* \textcolor{cyan}{\emph{Sub-task self-evolution}} */
            \STATE Generate $(1 - a_{(2),j}^t)$ offspring by $\mathcal{A}$, creating $\{V_{\text{se},j}^t\}$;
            \STATE /* \textcolor{cyan}{\emph{Cross-task knowledge transfer}} */
            \STATE Generate $a_{(2),j}^t$ offspring by $\mathcal{A}$, creating $\{V_{\text{kt},j}^t\}$; \COMMENT{source task from $a_{(1)}^t$, behaviors from $a_{(3)}^t$ (Sec.~\ref{sec:action})}
            \STATE Combine offspring: $\{V_j^t\}_{j=1}^{K} = \{V_{\text{se},j}^t\}_{j=1}^{K} \cup \{V_{\text{kt},j}^t\}_{j=1}^{K}$;

            \STATE /* \textcolor{cyan}{\emph{Evaluation and selection}} */
            \STATE Evaluate and obtain $\{V_{j}^t, f_j^t(V_{j}^t)\}_{j=1}^{K}$;
            \STATE Select next generation $\{X_j^{t+1}\}_{j=1}^{K}$ from parents $X_j^t$ and offspring $V_j^t$~(Sec.~\ref{sec:dynamics});
            
            \STATE /* \textcolor{cyan}{\emph{PPO training}} */
            \STATE Compute reward $\mathcal{R}^t$~(Eq.~\ref{eq:reward});
            \STATE Record transition $<s^t, a^t, \log\pi_\theta(a^t|s^t), \mathcal{R}^t, s^{t+1}>$;
            \IF{$t\ \%\ T_{\text{ppo}} = 0$}
                \STATE Compute policy gradients $\nabla_\theta (\mathcal{J})$ of $\mathcal{J}(\theta)$ in Eq.~\ref{eq:mdp};
                \STATE Update $\theta$ along $\nabla_\theta$ using multi-step PPO~\cite{refs:ppo}; 
            \ENDIF
            \STATE set $t = t + 1$;
        \ENDWHILE
    \ENDFOR
    \STATE set $l = l + 1$;
\ENDWHILE
\RETURN trained $\pi_{\theta}$
\end{algorithmic}
\end{algorithm}

\subsubsection{Training Procedure}\label{sec:ppo}
We now elaborate the meta-training and inference workflows of our \textbf{\texttt{MetaMTO}}. The pesudocode of the training is presented in \textbf{Algorithm~\ref{alg:training}}. The input includes i) an initialized meta-level policy parameter $\theta$; ii) a training problem distribution $\mathbb{D}_{\text{train}}$, which could be any subset of our AWCCI~(i.e., VS $\sim$ VL); iii) a low-level EMT framework, which in this paper we adopt a multi-population DE algorithm $\mathcal{A}$ with explicit knowledge transfer scheme~(similar with the algorithm structure in~\cite{refs:EBS,refs:MaTEA}); iv) the meta-level learning budget $L$ and the low-level optimization budget $G$. In each epoch $g$, we traverse each problem instance in $\mathbb{D}_{\text{train}}$, letting the meta-level policy $\theta$ interact with low-level EMT optimization process~(lines 4-16). The transitions are obtained through the interaction~(lines 17-18) and recorded, which is then used to compute policy gradients~(lines 20-21). In \textbf{\texttt{MetaMTO}}, we adopt the popular policy gradient method PPO~\cite{refs:ppo}. The learning occurs every $T_{\text{ppo}}$ optimization steps, and the parameters is updated~(line 22) for $k_{ppo}$ iterations in each learning step. For the inference, once the training completes, one can directly use the trained model to solve unseen MTO instances, the only difference is that in inference, actions are selected in a deterministic manner~(the one with maximal likelihood). We leave a more detailed settings at Sec.~\ref{sec:4.1}.  

\begin{table*}[t]
	\caption{Comparison in terms of in-distribution generalization performance.}
	\label{tab:comparitive_results}
	\centering
    \resizebox{0.92\textwidth}{!}{
	\begin{tabular}{cccccccc|c|c}\hline
	& \multicolumn{1}{c}{Ours}  & \multicolumn{2}{c}{L2T\cite{refs:L2T}} & \multicolumn{2}{c}{BLKT\cite{refs:BLKT}}                      & \multicolumn{2}{c|}{RLMFEA\cite{refs:RLMFEA}}                     & \multicolumn{2}{c}{MFEA\cite{refs:MFEA}}  \\ \hline
		problems & \multicolumn{1}{|c}{Avg obj} & \multicolumn{1}{|c}{Avg obj} & \multicolumn{1}{|c}{\textcolor{cyan}{+}/\textcolor{crimson}{-}} & \multicolumn{1}{|c}{Avg obj} & \multicolumn{1}{|c}{\textcolor{cyan}{+}/\textcolor{crimson}{-}} & \multicolumn{1}{|c}{Avg obj} & \multicolumn{1}{|c|}{\textcolor{cyan}{+}/\textcolor{crimson}{-}} & \multicolumn{1}{c|}{Avg obj} & \multicolumn{1}{c}{\textcolor{cyan}{+}/\textcolor{crimson}{-}} \\ \hline
		VS-1     & \multicolumn{1}{|c}{\cellcolor{gray!30}\textbf{0.1897}$\boldsymbol{\pm}$\textbf{0.0504}}& \multicolumn{1}{|c}{0.2983$\pm$0.0019} & \multicolumn{1}{|c}{\textcolor{cyan}{68}/\textcolor{crimson}{2}} & \multicolumn{1}{|c}{0.2740$\pm$0.0108} & \multicolumn{1}{|c}{\textcolor{cyan}{60}/\textcolor{crimson}{10}} & \multicolumn{1}{|c}{0.3613$\pm$0.0136} & \multicolumn{1}{|c|}{\textcolor{cyan}{68}/\textcolor{crimson}{2}} &     0.3688$\pm$0.0114       &   \textcolor{cyan}{68}/\textcolor{crimson}{2}                    \\ \hline
		VS-2     & \multicolumn{1}{|c}{\cellcolor{gray!30}\textbf{0.1083}$\boldsymbol{\pm}$\textbf{0.0119}} & \multicolumn{1}{|c}{0.2495$\pm$0.0021} & \multicolumn{1}{|c}{\textcolor{cyan}{183}/\textcolor{crimson}{27}} & \multicolumn{1}{|c}{0.2736$\pm$0.0118} & \multicolumn{1}{|c}{\textcolor{cyan}{182}/\textcolor{crimson}{28}} & \multicolumn{1}{|c}{0.3605$\pm$0.0082} & \multicolumn{1}{|c|}{\textcolor{cyan}{203}/\textcolor{crimson}{7}} &         0.3552$\pm$0.0063     &    \textcolor{cyan}{204}/\textcolor{crimson}{6}                   \\ \hline
		VS-3     & \multicolumn{1}{|c}{\cellcolor{gray!30}\textbf{0.0866}$\boldsymbol{\pm}$\textbf{0.0069}} & \multicolumn{1}{|c}{0.2225$\pm$0.0025} & \multicolumn{1}{|c}{\textcolor{cyan}{301}/\textcolor{crimson}{49}} & \multicolumn{1}{|c}{0.2487$\pm$0.0052} & \multicolumn{1}{|c}{\textcolor{cyan}{316}/\textcolor{crimson}{34}} & \multicolumn{1}{|c}{0.3146$\pm$0.0048} & \multicolumn{1}{|c|}{\textcolor{cyan}{350}/\textcolor{crimson}{0}} &         0.3140$\pm$0.0055    &     \textcolor{cyan}{349}/\textcolor{crimson}{1}                  \\ \hline
		VS-4     & \multicolumn{1}{|c}{\cellcolor{gray!30}\textbf{0.0865}$\boldsymbol{\pm}$\textbf{0.0031}} & \multicolumn{1}{|c}{0.2139$\pm$0.0026} & \multicolumn{1}{|c}{\textcolor{cyan}{255}/\textcolor{crimson}{95}} & \multicolumn{1}{|c}{0.2465$\pm$0.0043} & \multicolumn{1}{|c}{\textcolor{cyan}{329}/\textcolor{crimson}{21}} & \multicolumn{1}{|c}{0.3094$\pm$0.0055} & \multicolumn{1}{|c|}{\textcolor{cyan}{342}/\textcolor{crimson}{8}} &        0.3085$\pm$0.0042    &    \textcolor{cyan}{342}/\textcolor{crimson}{8}                   \\ \hline
		VS-5     & \multicolumn{1}{|c}{\cellcolor{gray!30}\textbf{0.0929}$\boldsymbol{\pm}$\textbf{0.0055}} & \multicolumn{1}{|c}{0.1557$\pm$0.0027} & \multicolumn{1}{|c}{\textcolor{cyan}{144}/\textcolor{crimson}{66}} & \multicolumn{1}{|c}{0.2079$\pm$0.0032} & \multicolumn{1}{|c}{\textcolor{cyan}{190}/\textcolor{crimson}{20}} & \multicolumn{1}{|c}{0.2750$\pm$0.0082} & \multicolumn{1}{|c|}{\textcolor{cyan}{210}/\textcolor{crimson}{0}} &         0.2788$\pm$0.0056     &    \textcolor{cyan}{208}/\textcolor{crimson}{2}                  \\ \hline
		VS-6     & \multicolumn{1}{|c}{\cellcolor{gray!30}\textbf{0.0959}$\boldsymbol{\pm}$\textbf{0.0143}} & \multicolumn{1}{|c}{0.2035$\pm$0.0072} & \multicolumn{1}{|c}{\textcolor{cyan}{60}/\textcolor{crimson}{10}} & \multicolumn{1}{|c}{0.2396$\pm$0.0092} & \multicolumn{1}{|c}{\textcolor{cyan}{66}/\textcolor{crimson}{4}} & \multicolumn{1}{|c}{0.3347$\pm$0.0149} & \multicolumn{1}{|c|}{\textcolor{cyan}{70}/\textcolor{crimson}{0}} &         0.3325$\pm$0.0189     &    \textcolor{cyan}{70}/\textcolor{crimson}{0}                   \\ \hline
		VS-7     & \multicolumn{1}{|c}{\cellcolor{gray!30}\textbf{0.1360}$\boldsymbol{\pm}$\textbf{0.0345}} & \multicolumn{1}{|c}{0.3154$\pm$0.0031} & \multicolumn{1}{|c}{\textcolor{cyan}{10}/\textcolor{crimson}{0}} & \multicolumn{1}{|c}{0.3550$\pm$0.0095} & \multicolumn{1}{|c}{\textcolor{cyan}{10}/\textcolor{crimson}{0}} & \multicolumn{1}{|c}{0.3120$\pm$0.0209} & \multicolumn{1}{|c|}{\textcolor{cyan}{10}/\textcolor{crimson}{0}} &         0.3173$\pm$0.0277     &    \textcolor{cyan}{10}/\textcolor{crimson}{0}                   \\ \hline
		S-1      & \multicolumn{1}{|c}{\cellcolor{gray!30}\textbf{0.1820}$\boldsymbol{\pm}$\textbf{0.0381}} & \multicolumn{1}{|c}{0.3074$\pm$0.0013} & \multicolumn{1}{|c}{\textcolor{cyan}{70}/\textcolor{crimson}{0}} & \multicolumn{1}{|c}{0.2744$\pm$0.0079} & \multicolumn{1}{|c}{\textcolor{cyan}{60}/\textcolor{crimson}{10}} & \multicolumn{1}{|c}{0.3769$\pm$0.0067} & \multicolumn{1}{|c|}{\textcolor{cyan}{69}/\textcolor{crimson}{1}} &     0.3716$\pm$0.0067          &    \textcolor{cyan}{68}/\textcolor{crimson}{2}                   \\ \hline
		S-2      & \multicolumn{1}{|c}{\cellcolor{gray!30}\textbf{0.1317}$\boldsymbol{\pm}$\textbf{0.0088}} & \multicolumn{1}{|c}{0.2694$\pm$0.0026} & \multicolumn{1}{|c}{\textcolor{cyan}{184}/\textcolor{crimson}{26}} & \multicolumn{1}{|c}{0.2787$\pm$0.0088} & \multicolumn{1}{|c}{\textcolor{cyan}{176}/\textcolor{crimson}{34}} & \multicolumn{1}{|c}{0.3707$\pm$0.0076} & \multicolumn{1}{|c|}{\textcolor{cyan}{203}/\textcolor{crimson}{7}} &     0.3697$\pm$0.0066              &      \textcolor{cyan}{202}/\textcolor{crimson}{8}                 \\ \hline
		S-3      & \multicolumn{1}{|c}{\cellcolor{gray!30}\textbf{0.1165}$\boldsymbol{\pm}$\textbf{0.0053}} & \multicolumn{1}{|c}{0.2434$\pm$0.0027} & \multicolumn{1}{|c}{\textcolor{cyan}{304}/\textcolor{crimson}{46}} & \multicolumn{1}{|c}{0.2531$\pm$0.0037} & \multicolumn{1}{|c}{\textcolor{cyan}{298}/\textcolor{crimson}{52}} & \multicolumn{1}{|c}{0.3287$\pm$0.0075} & \multicolumn{1}{|c|}{\textcolor{cyan}{348}/\textcolor{crimson}{2}} &     0.3270$\pm$0.0048             &      \textcolor{cyan}{348}/\textcolor{crimson}{2}                 \\ \hline
		S-4      & \multicolumn{1}{|c}{\cellcolor{gray!30}\textbf{0.1180}$\boldsymbol{\pm}$\textbf{0.0041}} & \multicolumn{1}{|c}{0.2371$\pm$0.0018} & \multicolumn{1}{|c}{\textcolor{cyan}{253}/\textcolor{crimson}{97}} & \multicolumn{1}{|c}{0.2518$\pm$0.0042} & \multicolumn{1}{|c}{\textcolor{cyan}{312}/\textcolor{crimson}{38}} & \multicolumn{1}{|c}{0.3247$\pm$0.0060} & \multicolumn{1}{|c|}{\textcolor{cyan}{345}/\textcolor{crimson}{5}} &     0.3250$\pm$0.0062            &     \textcolor{cyan}{345}/\textcolor{crimson}{5}                  \\ \hline
		S-5      & \multicolumn{1}{|c}{\cellcolor{gray!30}\textbf{0.1285}$\boldsymbol{\pm}$\textbf{0.0080}} & \multicolumn{1}{|c}{0.1823$\pm$0.0036} & \multicolumn{1}{|c}{\textcolor{cyan}{137}/\textcolor{crimson}{73}} & \multicolumn{1}{|c}{0.2108$\pm$0.0039} & \multicolumn{1}{|c}{\textcolor{cyan}{172}/\textcolor{crimson}{38}} & \multicolumn{1}{|c}{0.2952$\pm$0.0055} & \multicolumn{1}{|c|}{\textcolor{cyan}{210}/\textcolor{crimson}{0}} &     0.2953$\pm$0.0075             &       \textcolor{cyan}{210}/\textcolor{crimson}{0}                \\ \hline
		S-6      & \multicolumn{1}{|c}{\cellcolor{gray!30}\textbf{0.1266}$\boldsymbol{\pm}$\textbf{0.0149}} & \multicolumn{1}{|c}{0.2348$\pm$0.0065} & \multicolumn{1}{|c}{\textcolor{cyan}{60}/\textcolor{crimson}{10}} & \multicolumn{1}{|c}{0.2414$\pm$0.0045} & \multicolumn{1}{|c}{\textcolor{cyan}{65}/\textcolor{crimson}{5}} & \multicolumn{1}{|c}{0.3495$\pm$0.0140} & \multicolumn{1}{|c|}{\textcolor{cyan}{70}/\textcolor{crimson}{0}} &     0.3522$\pm$0.0191              &     \textcolor{cyan}{70}/\textcolor{crimson}{0}                  \\ \hline
		S-7      & \multicolumn{1}{|c}{\cellcolor{gray!30}\textbf{0.1596}$\boldsymbol{\pm}$\textbf{0.0131}} & \multicolumn{1}{|c}{0.3205$\pm$0.0025} & \multicolumn{1}{|c}{\textcolor{cyan}{10}/\textcolor{crimson}{0}} & \multicolumn{1}{|c}{0.3300$\pm$0.0430} & \multicolumn{1}{|c}{\textcolor{cyan}{10}/\textcolor{crimson}{0}} & \multicolumn{1}{|c}{ 0.3377$\pm$0.0256} & \multicolumn{1}{|c|}{\textcolor{cyan}{10}/\textcolor{crimson}{0}} &          0.3339$\pm$0.0194             &   \textcolor{cyan}{10}/\textcolor{crimson}{0}                     \\ \hline
		M-1      & \multicolumn{1}{|c}{\cellcolor{gray!30}\textbf{0.2605}$\boldsymbol{\pm}$\textbf{0.0303}} & \multicolumn{1}{|c}{0.3207$\pm$0.0015} & \multicolumn{1}{|c}{\textcolor{cyan}{70}/\textcolor{crimson}{0}} & \multicolumn{1}{|c}{0.2829$\pm$0.0091} & \multicolumn{1}{|c}{\textcolor{cyan}{62}/\textcolor{crimson}{8}} & \multicolumn{1}{|c}{0.3920$\pm$0.0101} & \multicolumn{1}{|c|}{\textcolor{cyan}{67}/\textcolor{crimson}{3}} &         0.3884$\pm$0.0049              &    \textcolor{cyan}{67}/\textcolor{crimson}{3}                   \\ \hline
		M-2      & \multicolumn{1}{|c}{\cellcolor{gray!30}\textbf{0.1838}$\boldsymbol{\pm}$\textbf{0.0080}} & \multicolumn{1}{|c}{0.3137$\pm$0.0016} & \multicolumn{1}{|c}{\textcolor{cyan}{186}/\textcolor{crimson}{24}} & \multicolumn{1}{|c}{0.2814$\pm$0.0093} & \multicolumn{1}{|c}{\textcolor{cyan}{165}/\textcolor{crimson}{45}} & \multicolumn{1}{|c}{0.4050$\pm$0.0058} & \multicolumn{1}{|c|}{\textcolor{cyan}{202}/\textcolor{crimson}{8}} &         0.3967$\pm$0.0073              &     \textcolor{cyan}{201}/\textcolor{crimson}{9}                  \\ \hline
		M-3      & \multicolumn{1}{|c}{\cellcolor{gray!30}\textbf{0.1713}$\boldsymbol{\pm}$\textbf{0.0058}} & \multicolumn{1}{|c}{0.2862$\pm$0.0018} & \multicolumn{1}{|c}{\textcolor{cyan}{299}/\textcolor{crimson}{51}} & \multicolumn{1}{|c}{0.2600$\pm$0.0049} & \multicolumn{1}{|c}{\textcolor{cyan}{256}/\textcolor{crimson}{94}} & \multicolumn{1}{|c}{0.3634$\pm$0.0041} & \multicolumn{1}{|c|}{\textcolor{cyan}{338}/\textcolor{crimson}{12}} &         0.3605$\pm$0.0051              &    \textcolor{cyan}{337}/\textcolor{crimson}{13}                   \\ \hline
		M-4      & \multicolumn{1}{|c}{\cellcolor{gray!30}\textbf{0.1749}$\boldsymbol{\pm}$\textbf{0.0050}} & \multicolumn{1}{|c}{0.2819$\pm$0.0019} & \multicolumn{1}{|c}{\textcolor{cyan}{286}/\textcolor{crimson}{64}} & \multicolumn{1}{|c}{0.2611$\pm$0.0038} & \multicolumn{1}{|c}{\textcolor{cyan}{276}/\textcolor{crimson}{74}} & \multicolumn{1}{|c}{0.3606$\pm$0.0045} & \multicolumn{1}{|c|}{\textcolor{cyan}{348}/\textcolor{crimson}{2}} &         0.3598$\pm$0.0051              &     \textcolor{cyan}{349}/\textcolor{crimson}{1}                  \\ \hline
		M-5      & \multicolumn{1}{|c}{\cellcolor{gray!30}\textbf{0.1659}$\boldsymbol{\pm}$\textbf{0.0062}} & \multicolumn{1}{|c}{0.2334$\pm$0.0024} & \multicolumn{1}{|c}{\textcolor{cyan}{169}/\textcolor{crimson}{41}} & \multicolumn{1}{|c}{0.2165$\pm$0.0037} & \multicolumn{1}{|c}{\textcolor{cyan}{151}/\textcolor{crimson}{59}} & \multicolumn{1}{|c}{0.3338$\pm$0.0089} & \multicolumn{1}{|c|}{\textcolor{cyan}{210}/\textcolor{crimson}{0}}   &  0.3317$\pm$0.0063                     &    \textcolor{cyan}{210}/\textcolor{crimson}{0}                   \\ \hline
		M-6      & \multicolumn{1}{|c}{\cellcolor{gray!30}\textbf{0.1957}$\boldsymbol{\pm}$\textbf{0.0153}} & \multicolumn{1}{|c}{0.2934$\pm$0.0043} & \multicolumn{1}{|c}{\textcolor{cyan}{59}/\textcolor{crimson}{11}} & \multicolumn{1}{|c}{0.2501$\pm$0.0045} & \multicolumn{1}{|c}{\textcolor{cyan}{53}/\textcolor{crimson}{17}} & \multicolumn{1}{|c}{0.3907$\pm$0.0125} & \multicolumn{1}{|c|}{\textcolor{cyan}{70}/\textcolor{crimson}{0}} &         0.3819$\pm$0.0151              &    \textcolor{cyan}{69}/\textcolor{crimson}{1}                   \\ \hline
		M-7      & \multicolumn{1}{|c}{\cellcolor{gray!30}\textbf{0.1813}$\boldsymbol{\pm}$\textbf{0.0210}} & \multicolumn{1}{|c}{0.3308$\pm$0.0042} & \multicolumn{1}{|c}{\textcolor{cyan}{10}/\textcolor{crimson}{0}} & \multicolumn{1}{|c}{0.3556$\pm$0.0053} & \multicolumn{1}{|c}{\textcolor{cyan}{10}/\textcolor{crimson}{0}} & \multicolumn{1}{|c}{0.3802$\pm$0.0325} & \multicolumn{1}{|c|}{\textcolor{cyan}{10}/\textcolor{crimson}{0}} &         0.3764$\pm$0.0322              &    \textcolor{cyan}{10}/\textcolor{crimson}{0}                   \\ \hline
		L-1      & \multicolumn{1}{|c}{\cellcolor{gray!30}\textbf{0.2687}$\boldsymbol{\pm}$\textbf{0.0215}} & \multicolumn{1}{|c}{0.3308$\pm$0.0017} & \multicolumn{1}{|c}{\textcolor{cyan}{70}/\textcolor{crimson}{0}} & \multicolumn{1}{|c}{0.2847$\pm$0.0131} & \multicolumn{1}{|c}{\textcolor{cyan}{57}/\textcolor{crimson}{13}} & \multicolumn{1}{|c}{0.4239$\pm$0.0055} & \multicolumn{1}{|c|}{\textcolor{cyan}{68}/\textcolor{crimson}{2}} &         0.4158$\pm$0.0045              &     \textcolor{cyan}{67}/\textcolor{crimson}{3}                  \\ \hline
		L-2      & \multicolumn{1}{|c}{\cellcolor{gray!30}\textbf{0.2195}$\boldsymbol{\pm}$\textbf{0.0071}} & \multicolumn{1}{|c}{0.3487$\pm$0.0021} & \multicolumn{1}{|c}{\textcolor{cyan}{199}/\textcolor{crimson}{11}} & \multicolumn{1}{|c}{0.2923$\pm$0.0064} & \multicolumn{1}{|c}{\textcolor{cyan}{141}/\textcolor{crimson}{69}} & \multicolumn{1}{|c}{0.4346$\pm$0.0033} & \multicolumn{1}{|c|}{\textcolor{cyan}{202}/\textcolor{crimson}{8}} &         0.4335$\pm$0.0056              &    \textcolor{cyan}{201}/\textcolor{crimson}{9}                   \\ \hline
		L-3      & \multicolumn{1}{|c}{\cellcolor{gray!30}\textbf{0.2121}$\boldsymbol{\pm}$\textbf{0.0039}} & \multicolumn{1}{|c}{0.3201$\pm$0.0021} & \multicolumn{1}{|c}{\textcolor{cyan}{309}/\textcolor{crimson}{41}} & \multicolumn{1}{|c}{0.2669$\pm$0.0032} & \multicolumn{1}{|c}{\textcolor{cyan}{205}/\textcolor{crimson}{145}} & \multicolumn{1}{|c}{0.4020$\pm$0.0040} & \multicolumn{1}{|c|}{\textcolor{cyan}{339}/\textcolor{crimson}{11}} &         0.3997$\pm$0.0029              &   \textcolor{cyan}{338}/\textcolor{crimson}{12}                    \\ \hline
		L-4      & \multicolumn{1}{|c}{\cellcolor{gray!30}\textbf{0.2167}$\boldsymbol{\pm}$\textbf{0.0017}} & \multicolumn{1}{|c}{0.3128$\pm$0.0024} & \multicolumn{1}{|c}{\textcolor{cyan}{298}/\textcolor{crimson}{52}} & \multicolumn{1}{|c}{0.2663$\pm$0.0026} & \multicolumn{1}{|c}{\textcolor{cyan}{232}/\textcolor{crimson}{118}} & \multicolumn{1}{|c}{0.4006$\pm$0.0031} & \multicolumn{1}{|c|}{\textcolor{cyan}{349}/\textcolor{crimson}{1}} &         0.3961$\pm$0.0036              &   \textcolor{cyan}{349}/\textcolor{crimson}{1}                    \\ \hline
		L-5      & \multicolumn{1}{|c}{\cellcolor{gray!30}\textbf{0.1993}$\boldsymbol{\pm}$\textbf{0.0038}} & \multicolumn{1}{|c}{0.2701$\pm$0.0022} & \multicolumn{1}{|c}{\textcolor{cyan}{169}/\textcolor{crimson}{41}} & \multicolumn{1}{|c}{0.224$\pm$(0.0056} & \multicolumn{1}{|c}{\textcolor{cyan}{112}/\textcolor{crimson}{98}} & \multicolumn{1}{|c}{0.3652$\pm$0.0053} & \multicolumn{1}{|c|}{\textcolor{cyan}{210}/\textcolor{crimson}{0}} &         0.3695$\pm$0.0026              &     \textcolor{cyan}{210}/\textcolor{crimson}{0}                  \\ \hline
		L-6      & \multicolumn{1}{|c}{\cellcolor{gray!30}\textbf{0.2384}$\boldsymbol{\pm}$\textbf{0.0066}} & \multicolumn{1}{|c}{0.3302$\pm$0.0043} & \multicolumn{1}{|c}{\textcolor{cyan}{68}/\textcolor{crimson}{2}} & \multicolumn{1}{|c}{0.2540$\pm$0.0077} & \multicolumn{1}{|c}{\textcolor{cyan}{46}/\textcolor{crimson}{24}} & \multicolumn{1}{|c}{0.4162$\pm$0.0108} & \multicolumn{1}{|c|}{\textcolor{cyan}{70}/\textcolor{crimson}{0}} &         0.4236$\pm$0.0113              &     \textcolor{cyan}{69}/\textcolor{crimson}{1}                  \\ \hline
		L-7      & \multicolumn{1}{|c}{\cellcolor{gray!30}\textbf{0.2040}$\boldsymbol{\pm}$\textbf{0.0168}} & \multicolumn{1}{|c}{0.3460$\pm$0.0073} & \multicolumn{1}{|c}{\textcolor{cyan}{10}/\textcolor{crimson}{0}} & \multicolumn{1}{|c}{0.3628$\pm$0.0104} & \multicolumn{1}{|c}{\textcolor{cyan}{10}/\textcolor{crimson}{0}} & \multicolumn{1}{|c}{0.4091$\pm$0.0139} & \multicolumn{1}{|c|}{\textcolor{cyan}{10}/\textcolor{crimson}{0}} &         0.4168$\pm$0.0416              &    \textcolor{cyan}{10}/\textcolor{crimson}{0}                   \\ \hline
		VL-1     & \multicolumn{1}{|c}{\cellcolor{gray!30}\textbf{0.2736}$\boldsymbol{\pm}$\textbf{0.0197}} & \multicolumn{1}{|c}{0.3411$\pm$0.0018} & \multicolumn{1}{|c}{\textcolor{cyan}{70}/\textcolor{crimson}{0}} & \multicolumn{1}{|c}{0.3018$\pm$0.0079} & \multicolumn{1}{|c}{\textcolor{cyan}{56}/\textcolor{crimson}{14}} & \multicolumn{1}{|c}{0.4269$\pm$0.0050} & \multicolumn{1}{|c|}{\textcolor{cyan}{68}/\textcolor{crimson}{2}} &         0.4316$\pm$0.0096              &   \textcolor{cyan}{70}/\textcolor{crimson}{0}                    \\ \hline
		VL-2     & \multicolumn{1}{|c}{\cellcolor{gray!30}\textbf{0.2415}$\boldsymbol{\pm}$\textbf{0.0065}} & \multicolumn{1}{|c}{0.3656$\pm$0.0018} & \multicolumn{1}{|c}{\textcolor{cyan}{201}/\textcolor{crimson}{9}} & \multicolumn{1}{|c}{0.3255$\pm$0.0062} & \multicolumn{1}{|c}{\textcolor{cyan}{163}/\textcolor{crimson}{47}} & \multicolumn{1}{|c}{0.4498$\pm$0.0043} & \multicolumn{1}{|c|}{\textcolor{cyan}{198}/\textcolor{crimson}{12}} &         0.4479$\pm$0.0033              &     \textcolor{cyan}{198}/\textcolor{crimson}{12}                  \\ \hline
		VL-3     & \multicolumn{1}{|c}{\cellcolor{gray!30}\textbf{0.2413}$\boldsymbol{\pm}$\textbf{0.0054}} & \multicolumn{1}{|c}{0.3370$\pm$0.0007} & \multicolumn{1}{|c}{\textcolor{cyan}{321}/\textcolor{crimson}{29}} & \multicolumn{1}{|c}{0.2943$\pm$0.0036} & \multicolumn{1}{|c}{\textcolor{cyan}{248}/\textcolor{crimson}{102}} & \multicolumn{1}{|c}{0.4145$\pm$0.0026} & \multicolumn{1}{|c|}{\textcolor{cyan}{338}/\textcolor{crimson}{12}} &         0.4168$\pm$0.0030              &      \textcolor{cyan}{339}/\textcolor{crimson}{11}                 \\ \hline
		VL-4     & \multicolumn{1}{|c}{\cellcolor{gray!30}\textbf{0.2454}$\boldsymbol{\pm}$\textbf{0.0044}} & \multicolumn{1}{|c}{0.3335$\pm$0.0018} & \multicolumn{1}{|c}{\textcolor{cyan}{323}/\textcolor{crimson}{27}} & \multicolumn{1}{|c}{0.2919$\pm$0.0033} & \multicolumn{1}{|c}{\textcolor{cyan}{267}/\textcolor{crimson}{83}} & \multicolumn{1}{|c}{0.4174$\pm$0.0028} & \multicolumn{1}{|c|}{\textcolor{cyan}{350}/\textcolor{crimson}{0}} &         0.4168$\pm$0.0032              &   \textcolor{cyan}{349}/\textcolor{crimson}{1}                    \\ \hline
		VL-5     & \multicolumn{1}{|c}{\cellcolor{gray!30}\textbf{0.2268}$\boldsymbol{\pm}$\textbf{0.0047}} & \multicolumn{1}{|c}{0.2881$\pm$0.0011} & \multicolumn{1}{|c}{\textcolor{cyan}{189}/\textcolor{crimson}{21}} & \multicolumn{1}{|c}{0.2550$\pm$0.0043} & \multicolumn{1}{|c}{\textcolor{cyan}{141}/\textcolor{crimson}{69}} & \multicolumn{1}{|c}{0.3878$\pm$0.0045} & \multicolumn{1}{|c|}{\textcolor{cyan}{210}/\textcolor{crimson}{0}} &         0.3854$\pm$0.0030              &    \textcolor{cyan}{210}/\textcolor{crimson}{0}                  \\ \hline
		VL-6     & \multicolumn{1}{|c}{\cellcolor{gray!30}\textbf{0.2676}$\boldsymbol{\pm}$\textbf{0.0103}} & \multicolumn{1}{|c}{0.3483$\pm$0.0023} & \multicolumn{1}{|c}{\textcolor{cyan}{69}/\textcolor{crimson}{1}} & \multicolumn{1}{|c}{0.2956$\pm$0.0104} & \multicolumn{1}{|c}{\textcolor{cyan}{54}/\textcolor{crimson}{16}} & \multicolumn{1}{|c}{0.4361$\pm$0.0080} & \multicolumn{1}{|c|}{\textcolor{cyan}{69}/\textcolor{crimson}{1}} &         0.4339$\pm$0.0090              &      \textcolor{cyan}{70}/\textcolor{crimson}{0}                 \\ \hline
		VL-7     & \multicolumn{1}{|c}{\cellcolor{gray!30}\textbf{0.2349}$\boldsymbol{\pm}$\textbf{0.0251}} & \multicolumn{1}{|c}{0.3923$\pm$0.0060} & \multicolumn{1}{|c}{\textcolor{cyan}{10}/\textcolor{crimson}{0}} & \multicolumn{1}{|c}{0.3617$\pm$0.0083} & \multicolumn{1}{|c}{\textcolor{cyan}{10}/\textcolor{crimson}{0}} & \multicolumn{1}{|c}{0.4400$\pm$0.0184} & \multicolumn{1}{|c|}{\textcolor{cyan}{10}/\textcolor{crimson}{0}} &         0.4388$\pm$0.0343              &     \textcolor{cyan}{10}/\textcolor{crimson}{0}                  \\ \hline
		total         & \multicolumn{1}{|c}{\cellcolor{gray!30}\textbf{0.1823}$\boldsymbol{\pm}$\textbf{0.0128}} & \multicolumn{1}{|c}{0.2908$\pm$0.0028} & \multicolumn{1}{|c}{\textcolor{cyan}{5424}/\textcolor{crimson}{926}} & \multicolumn{1}{|c}{0.2763$\pm$0.0075} & \multicolumn{1}{|c}{\textcolor{cyan}{4961}/\textcolor{crimson}{1389}} & \multicolumn{1}{|c}{0.3743$\pm$0.0091} & \multicolumn{1}{|c|}{\textcolor{cyan}{6244}/\textcolor{crimson}{106}} &    0.3735$\pm$0.0104            &    \textcolor{cyan}{6237}/\textcolor{crimson}{113}                   \\ \hline
	\end{tabular}
    }
\end{table*}

\begin{figure*}[t]
\centering
\includegraphics[width=0.99\textwidth]{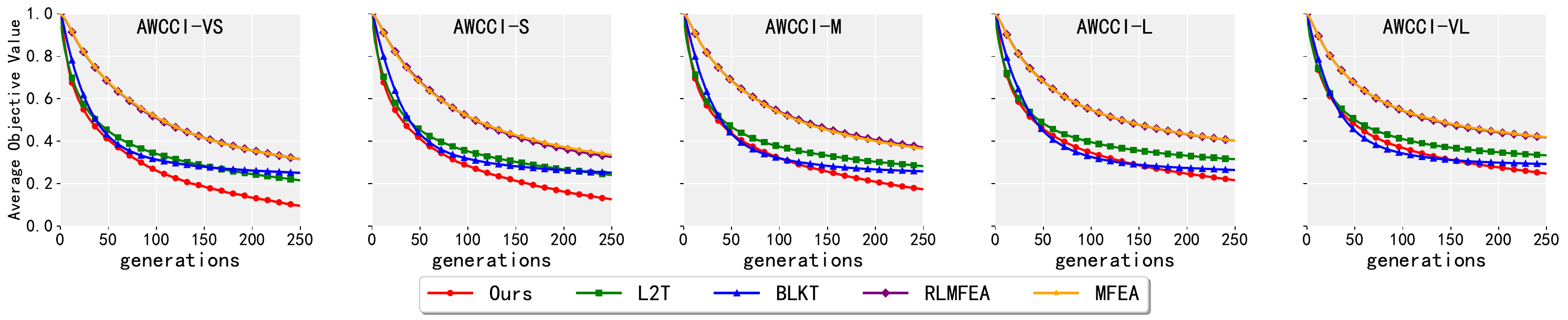}
\caption{Convergence curves on the five augmented test sets AWCCI-VS to AWCCI-VL.}
\label{fig:convergence_curves}
\end{figure*}

\section{Results and Discussions}\label{sec:4}
In  this section, we conduct meticulous experimental validation to demonstrate the effectiveness and performance advantages of our approach. Specifically, we delve into following key research questions: \textbf{RQ \#1:} have \textbf{\texttt{MetaMTO}} learned systematic and generalizable knowledge transfer policy for diverse MTO distributions? \textbf{RQ \#2:} can the learned policy perform robustly under extreme distribution shifts? \textbf{RQ \#3:} what factors contribute to the superior performance of \textbf{\texttt{MetaMTO}} and can these factors be interpreted  intuitively?  \textbf{RQ \#4:} are our designs for learning where, what and how to transfer knowledge really play important roles? Below we first introduce the experimental settings and then discuss these RQs respectively. To ensure the reproducibility, we provide sourcecodes of our project at \url{https://anonymous.4open.science/r/MetaMTO-B634}.

\subsection{Experimental Setup}\label{sec:4.1}

\textbf{MetaMTO:} The low-level EMT framework is a multi-population DE that is instantiated with population size as 50 for each sub-task and with the default optimization budget $G=250$ generations. For each sub-task's self-evolution routine, the setting adopts DE/rand/1 and binomial crossover, with $F=0.5$ and $Cr=0.7$. The other parts in this EMT framework are all controlled by our proposed multi-role RL system. For the PPO we used to train our RL system, its $T_{ppo}$ is set as 10 and $k_{ppo}$ is set as 3, other hyper-parameters such as the epsilon-greedy rate $\epsilon$, learning rate $\eta$ and discount factor $\gamma$ are set to 0.2, 0.0003 and 0.99 respectively. The training lasts $L=10$ epochs.

\textbf{Baselines:} We select four representative EMT approaches. MFEA~\cite{refs:MFEA} and BLKT~\cite{refs:BLKT} represent strong EMT baselines for implicit and explicit knowledge transfer paradigms, respectively. We also include two MetaBBO baselines for direct comparison to our approach: RLMFEA~\cite{refs:RLMFEA} leverages q-learning to train a parameter control policy for MFEA; L2T~\cite{refs:L2T} is proposed recently  to control knowledge transfer in the explicit EMT framework. We adopt the settings in their original papers. Note that L2T is originally proposed to address 2-task optimization problem, while the testbed in our experiment is AWCCI, of which the instances are 10-task problems. We hence revise the MLP architecture in L2T to make it compatible with our settings. 

\textbf{Train \& Test Settings:} We have three MetaBBO approaches~(i.e., RLMFEA, L2T and our \textbf{\texttt{MetaMTO}}) in the experiments. They are trained using our proposed AWCCI problem set. More specifically, we first set a training random seed and then generate five problem sets~(i.e., AWCCI-VS $\sim$ AWCCI-VL) according to Eq.~\ref{eq:awcci}. Then for each approach, we train its policy model exclusively on each of the five problem sets, which results in five trained models. During the testing phase, we first set a testing random seed and then generate five new problem sets correspondingly. The five trained models of a MetaBBO approach are then tested on corresponding problem set. For example, a model trained on AWCCI-VS training set is tested on the corresponding AWCCI-VS testing set. This testing protocol ensure a fine-grained comparison among the approaches across different MTO distributions. It is important to note that the protocol described above is designed for our in-distribution tests. In addition to this, we conduct a series of out-of-distribution experiments to assess model robustness, including evaluations across different difficulty distributions (e.g., training on one level and testing on another) and on problems with a varying number of tasks, as will be detailed in subsequent sections. Besides, MFEA and BLKT are directly tested on all testing sets. Each approach is tested for $N=10$ independent runs to reduce experimental variance.

\textbf{Performance Metrics:} To objectively reflect the performance difference between EMT approaches, two issues have to be addressed: i) a MTO problem include multiple sub-tasks, which anticipates proper aggregation of performance statistics; ii) the multiple sub-tasks may hold various objective value scales, which require careful normalization tricks. To this end, we propose a normalized performance metric~(smaller is better) to measure the performance of an approach $\mathcal{A}$ on an MTO instance $\mathcal{I}$ with $K$ sub-tasks:
\begin{equation}
\begin{aligned}
    \text{perf}(\mathcal{A},\mathcal{I}) = \frac{1}{K}\sum_{j=1}^{K}\text{perf}_j,\\
    \text{perf}_j = \frac{1}{N}\sum_{i=1}^{N} \frac{f_{j,i}^G - f_j^*}{f_{j,i}^0 - f_j^*}.
\end{aligned}    
\end{equation}
where $\text{perf}_j$ is the normalized performance of $\mathcal{A}$ on $j$-th sub-task across the $N=10$ independent test runs, $f_{j,i}^G$ denotes the optimal objective value found in $G$ generations, $f_{j,i}^0$ denotes the optimal in the initial population, $f_j^*$ denotes the true~(proxy\footnote{A proxy optimal can be easily obtained by traversing the optimal values found by all baselines and across all test runs.}) optimal of this sub-task.  

\subsection{Performance Comparison}
\subsubsection{In-distribution Generalization (RQ \#1)}
In this part, we focus on the in-distribution generalization ability of different methods. For each of the augmented test sets AWCCI-VS to AWCCI-VL, we test all approaches on the 127 MTO instance for 10 independent runs. In Table~\ref{tab:comparitive_results}, we report the averaged $\text{perf}(\mathcal{A},\mathcal{I})$ across MTO instances in each combination-based sub-set. For example, VS-4 denotes the MTO instances that include 4 of 7 different basic functions in its 10 sub-tasks. Hence, VS-4 include $C_7^4=35$ instances. We also attached significance statistics beside the normalized performance of four baselines to tell how many test runs our \textbf{\texttt{MetaMTO}} performs significantly better or worse than the others. The last row of the table is the summary across all $6350$ ($127\times5\times10$) test runs. 

As shown in Table \ref{tab:comparitive_results}, our approach significantly outperforms all four baselines across all test subsets in terms of the average normalized fitness and the number of winning problem instances. The superior performance validates the effectiveness of our proposed learning-based method, which automatically and comprehensively resolves the three key challenges of where to transfer, what to transfer and how to transfer. To be more specific, among the two explicit EMT algorithms, L2T outperforms BLKT on the VS and S subsets. However, its performance declines gradually on the M, L, and VL subsets, where it is surpassed by BLKT. In terms of the overall average performance, BLKT achieves superior results compared to L2T. The possible reason is that the block-based knowledge transfer mechanism in BLKT exhibits stronger optimization capability on more complex problems. Whereas, L2T suffers from limited generalization ability, which leads to a rapid degradation in solution quality. Further, in the comparison between the two implicit EMT algorithms, RLMFEA and MFEA demonstrate comparable optimization ability and RLMFEA is marginally inferior to MFEA. A possible explanation is that the learning-based knowledge transfer probability mechanism in RLMFEA lacks sufficient generalization capability, which limits its ability to effectively enhance problem solving on the AWCCI dataset. When considering both implicit and explicit EMT algorithms, RLMFEA and MFEA demonstrate notably inferior performance compared to BLKT and L2T. A likely reason is that the implicit methods follow the conventional MFEA mechanism and framework and lag behind the novel transfer strategies employed by explicit methods BLKT and L2T. Apparently, The latter ones  demonstrates superior problem-solving capability. Additionally, to further present the convergence behavior of our approach and the four compared algorithms, we illustrate the convergence curves for all subsets of the test set in the Fig. \ref{fig:convergence_curves}. It can be observed that the convergence curves cross-validate our observation in Table~\ref{tab:comparitive_results}: our approach consistently converges to better objective values than the baselines.

\begin{figure}[t]
\centering
\includegraphics[width=0.99\columnwidth]{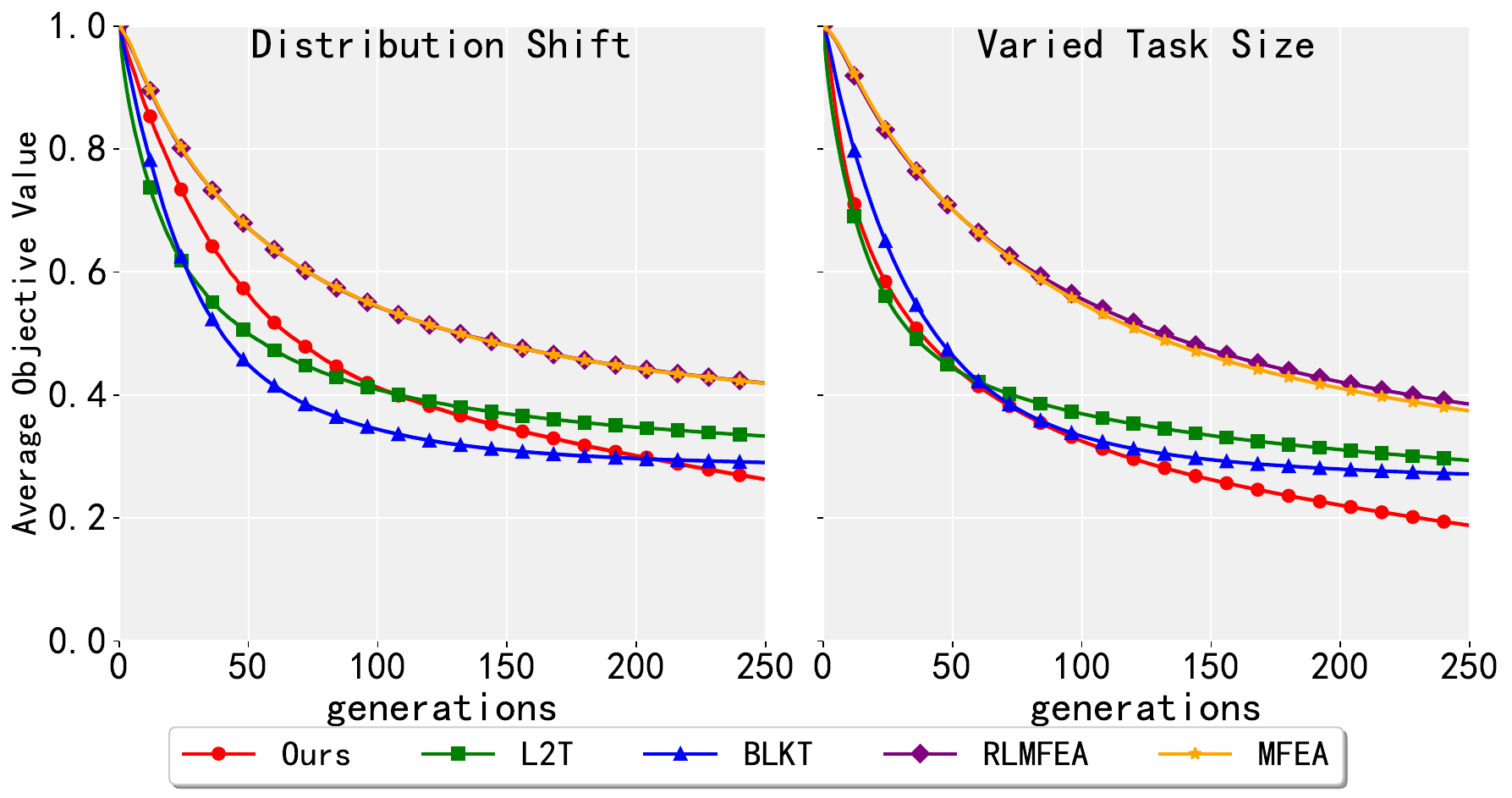}
\caption{Left is the comparison by shifting from training on AWCCI-VS to testing on AWCCI-VL, and right is the comparison by shifting from training on 10 sub-tasks to testing on 30 sub-tasks.}
\label{fig:generalization}
\end{figure}
\subsubsection{Out-of-distribution Generalization (RQ \#2)}
To validate the ture generalization potential of our \textbf{\texttt{MetaMTO}}, we conduct two out-of-distribution generalization tests on all approaches. Specifically, we focus on scenarios which feature extreme distribution shifts and where the number of sub-tasks varies. The former case occurs when an online computing server encounter high throughput optimization tasks from different user distributions. The latter case occurs when the server encounters unstable optimization request stream. We provide the following discussion on these two aspects. 

\textbf{Distribution Shift:} We pay attention to an extreme distribution shift case, where the approaches trained on AWCCI-VS are directly used to optimize instances in AWCCI-VL. We plot the convergence curves of all approaches in the left of Fig.~\ref{fig:generalization}~(averaged normalized objective across all tested instances and 10 independent runs). We can observe that our approach shows robust optimization performance under such severe distribution shift. In contrast, though the recent L2T framework performs with certain performance advantage in AWCCI-VS test set~(in-distribution) against human-crafted baseline such as BLKT, its generalization towards unseen distributional characteristics is limited and hence underperforms BLKT in the figure. The overall results demonstrate that our \textbf{\texttt{MetaMTO}} is more capable of dealing with multitask scenario in the wild.

\textbf{Varied Task Size:} We next explore whether \textbf{\texttt{MetaMTO}} could provide stable optimization when the number of sub-tasks in a MTO instance varies. Specifically, since our proposed neural network architecture uses attention-based blocks, \textbf{\texttt{MetaMTO}} is capable of processing MTO problems with different task sizes. In this test, we use approaches trained on AWCCI-VS to optimize a newly constructed AWCCI-VS-30 test set, instances of which comprises 30 sub-tasks~(larger than AWCCI-VS which comprises 10 sub-tasks per instance). We plot the convergence curves of all approaches in the right of Fig.~\ref{fig:generalization}~(averaged normalized objective across all tested instances and 10 independent runs). The results present a clear conclusion that our approach, once trained, could be used at unstable multitask environment.

\begin{figure}[t]
\centering
\includegraphics[width=0.99\columnwidth]{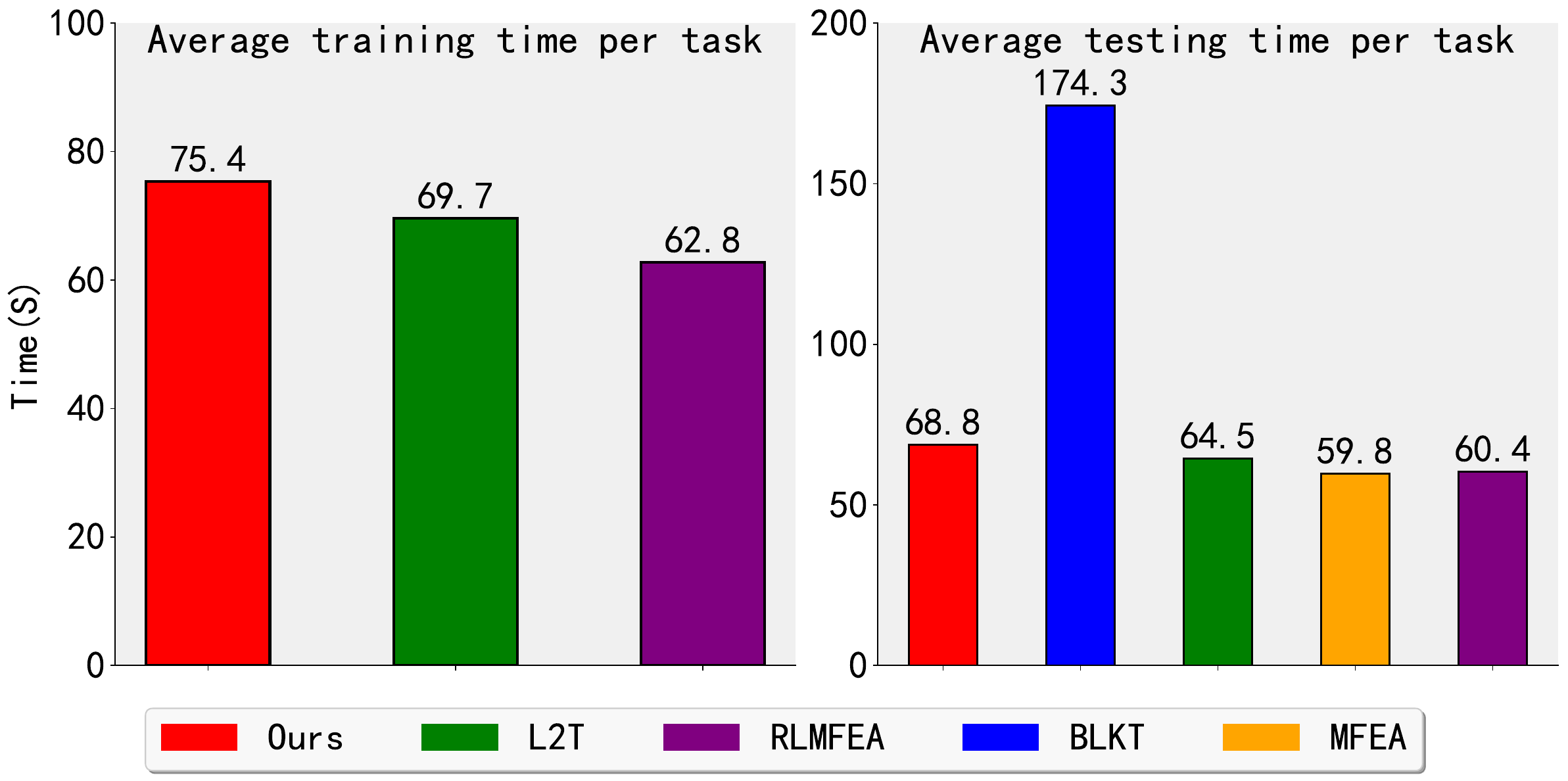}
\caption{Left is the average training time and right is the testing time per task.}
\label{fig:train_test_time}
\end{figure}

\begin{table}[t]
    \caption{The proportion of time between network inference and algorithmic problem solving}
    \label{tab:time_ratio}	
	\centering
    \resizebox{0.75\columnwidth}{!}{
	\begin{tabular}{c c}
    \toprule[2pt]
	\textsc{network inference} & \textsc{algorithm solving}  \\ 
    \midrule
	2.56\% & 97.44\% \\ 
    \bottomrule[2pt]
	\end{tabular}
    }    
\end{table}

\subsubsection{Runtime Complexity}
As our approach is learning-based, its computational efficiency is therefore a critical concern. To this end, we compare the average training time of our approach with RLMFEA and L2T and compare the average testing time against all the four compared algorithms in Fig.~\ref{fig:train_test_time}. Additionally, the overall time proportion between network inference and algorithmic problem solving is presented in Table~\ref{tab:time_ratio}. The results show that our approach remains the same order of average training time with the other two learning-based baselines. For inference, BLKT requires significantly more computational resources to achieve its state-of-the-art performance. This can be attributed to several computationally expensive operations in its optimization process including dividing individuals into blocks, performing block-based k-means clustering and reconstructing individuals from blocks. Our approach consumes the second in test time, but the gap with the other baselines is still acceptable. As further evidenced by Table \ref{tab:time_ratio}, the network inference time constitutes only a small fraction of the total runtime. This reveals the fact that our proposed approach improves performance with negligible computational complexity. In short, our approach achieves better solution quality without substantial time increase, making it both efficient and effective.

\begin{figure*}[t]
\centering
\includegraphics[width=0.99\textwidth]{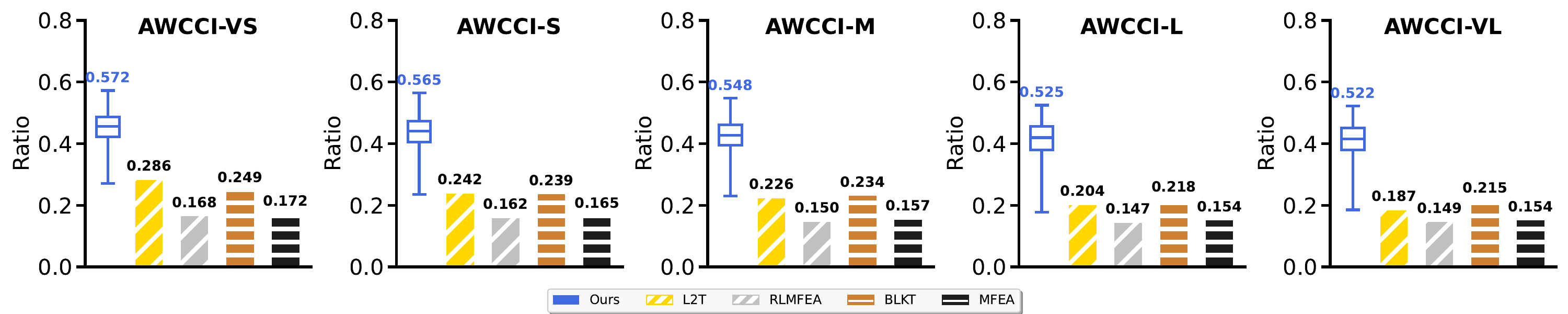}
\caption{Average KT success ratio over the five subsets}
\label{fig:KT_success_ratio}
\end{figure*}

\subsection{Attribution Analysis (RQ \#3)}
\subsubsection{Successful Knowledge Transfer}
An intuitive factor that impacts EMT approaches is the effectiveness of knowledge transfer. Given the sub-tasks of a MTO problem at hand, positive knowledge transfer could share useful genetic information between sub-tasks hence accelerates convergence, while negative knowledge transfer may harms the co-evolution among them. To this end, we summarize knowledge transfer success rates of different methods, which is defined as the proportion of survived transferred solutions, averaged across the 250 generations of all test runs. A higher knowledge transfer success rate indicates that an approach can effectively recognize similarity between sub-tasks and hence capable of promoting positive knowledge transfer. This metric indirectly reflects the effectiveness of the knowledge transfer mechanism, which helps interpret superiority of an approach. The results on the five test sets AWCCI-VS to AWCCI-VL are presented in Fig. \ref{fig:KT_success_ratio} from left to right respectively. These results provide us a clear observation that, by using our proposed multi-role RL system to learn systematic and fine-grained knowledge transfer, a basic multi-population DE backbone can be boosted with more accurate positive knowledge transfer and hence surpasses advanced human-crafted or learning-assisted baselines.

\subsubsection{Successful Task Routing}
\begin{figure}[t]
\centering
\includegraphics[width=0.99\columnwidth]{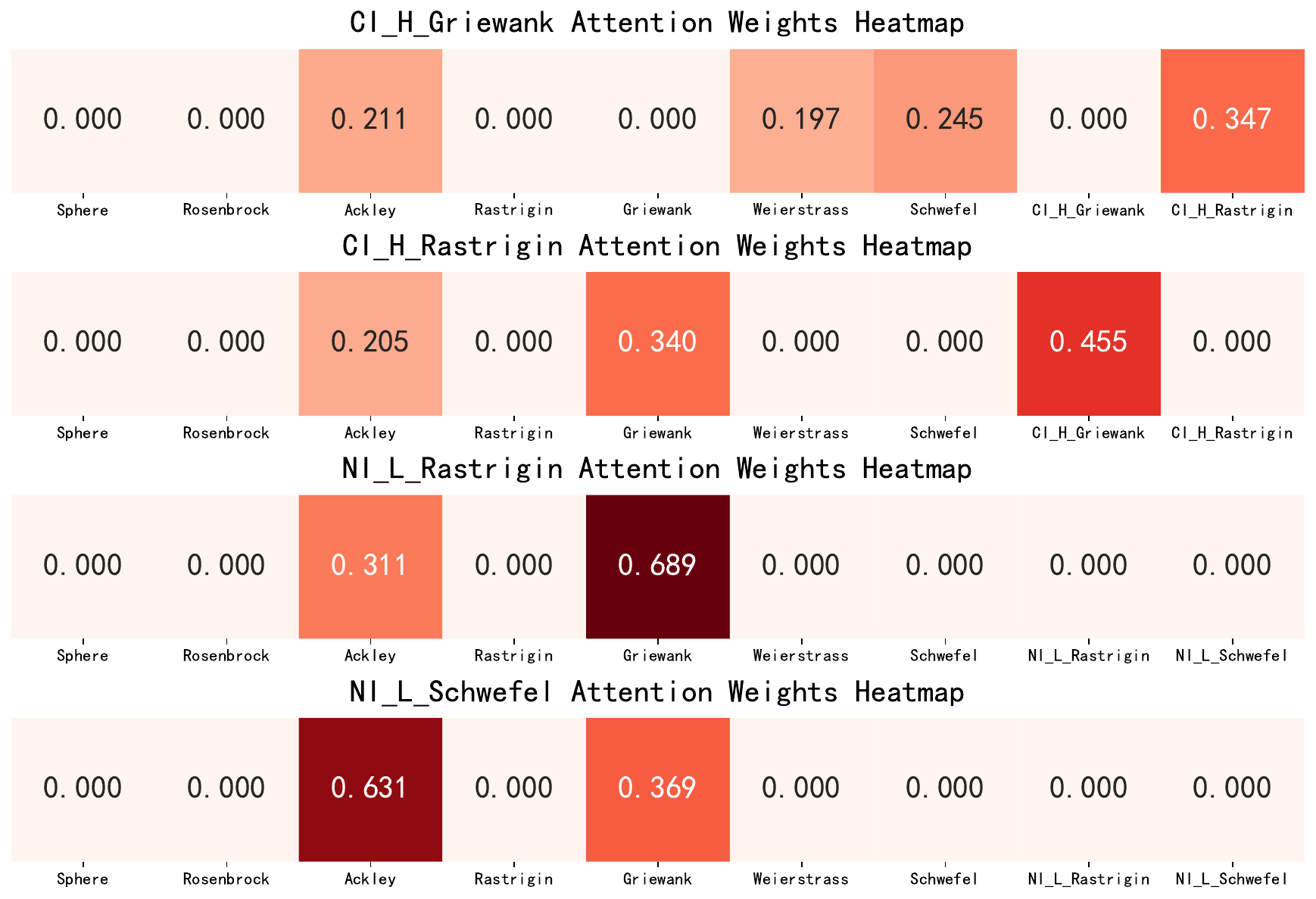}
\caption{Inter-task similarity recognition capability of our approach.}
\label{fig:attention_heatmaps}
\end{figure}

Given the successful knowledge transfer observed above, we further provide an in-depth analysis on the rationale behind such positive knowledge transfer of our \textbf{\texttt{MetaMTO}}. This analysis benefits from our attention-based neural network design, where the TR agent in the proposed multi-role RL system compute similarity scores via its attention layers. The objective is to discover whether this mechanism demonstrates the capability to identify the relevant and complementary task while filter out irrelevant ones. To create controlled test cases, we leverage two task pairs from the CEC2017 multitask optimization dataset~\cite{refs:CEC2017_WCCI2020}, whose inter-task relationships are known: 1) the CI\_H (Complete Intersection and High Similarity) instance, which consists of two sub-tasks CI\_H\_Griewank and CI\_H\_Rastrigin; 2) the NI\_L (No Intersection and Low Similarity) instance, which  consists of two sub-tasks NI\_L\_Rastrigin and NI\_L\_Schwefel. The former case has been validated as similar sub-tasks that knowledge transfer between them is preferable, while the latter case has been validated as irrelevant sub-tasks that should not have knowledge transfer. We then constructed two special, 9-task MTO problems: the first 7 tasks are selected from the WCCI2020~\cite{WCCI2020}, each uses a different basic function; and the remaining two tasks are from either the CI\_H or NI\_L pairs. We then use the model trained on AWCCI-S to optimize these two 9-task instances. We show the average attention scores output by the TR agent in the trained model across all test runs and all 250 optimization generations in Fig.~\ref{fig:attention_heatmaps}. It can be observed that the attention scores between the two CI\_H tasks are mutually the highest, indicating that the model consistently  identifies them as each other's most valuable source for knowledge transfer. Conversely, the attention scores between the two NI\_L tasks are the lowest, suggesting that neither of them ever selects the other as source knowledge. These results provide direct evidence that the successful knowledge transfer of our approach is rooted in its ability to perform intelligent task routing, driven by an accurate, learned recognition of inter-task similarity.

\subsection{Ablation Studies (RQ \#4)}
\begin{figure}[t]
    \centering
    \subfloat[AWCCI-VS]{\includegraphics[width=0.95\columnwidth]{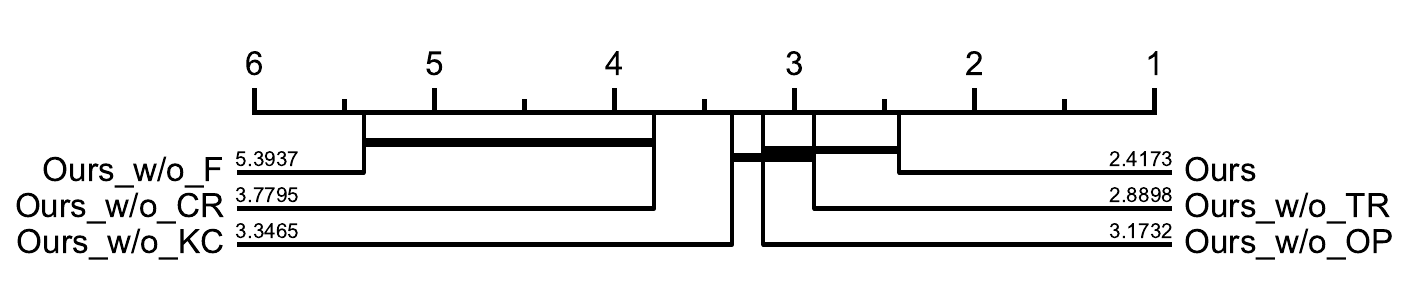}
    \label{subfig:cd_vs}} \\
    \subfloat[AWCCI-S]{\includegraphics[width=0.95\columnwidth]{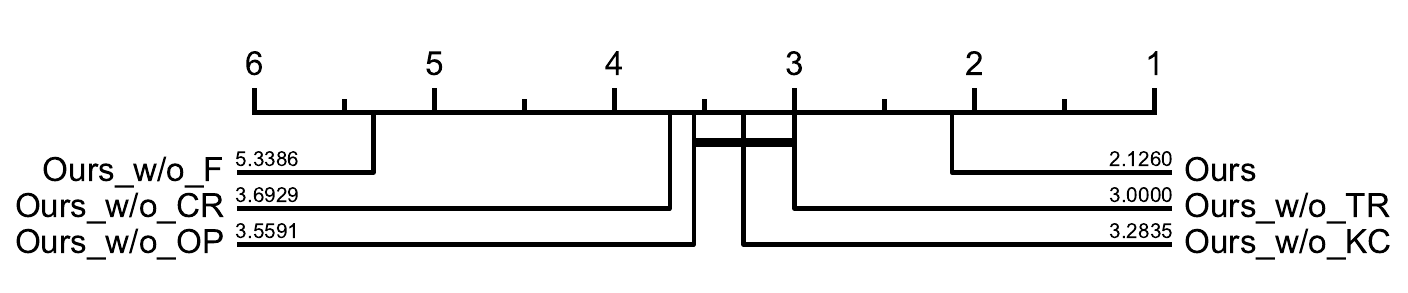}
    \label{subfig:cd_s}} \\

    \subfloat[AWCCI-M]{\includegraphics[width=0.95\columnwidth]{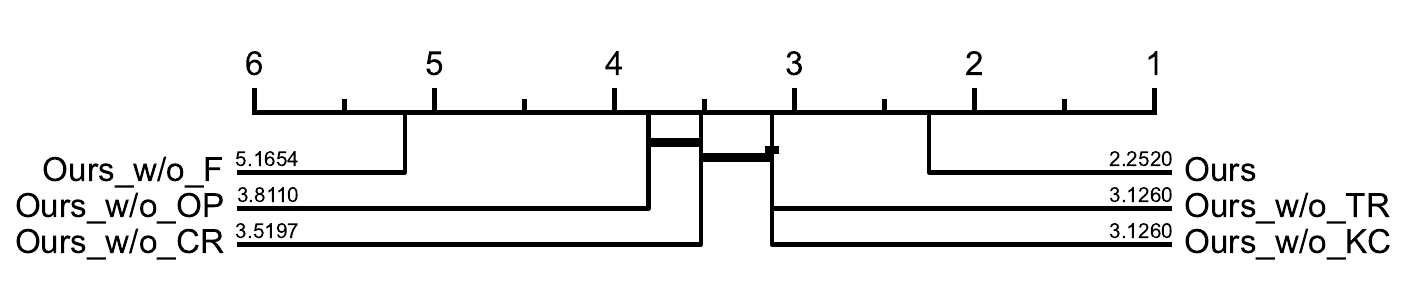}
    \label{subfig:cd_m}} \\
    \subfloat[AWCCI-L]{\includegraphics[width=0.95\columnwidth]{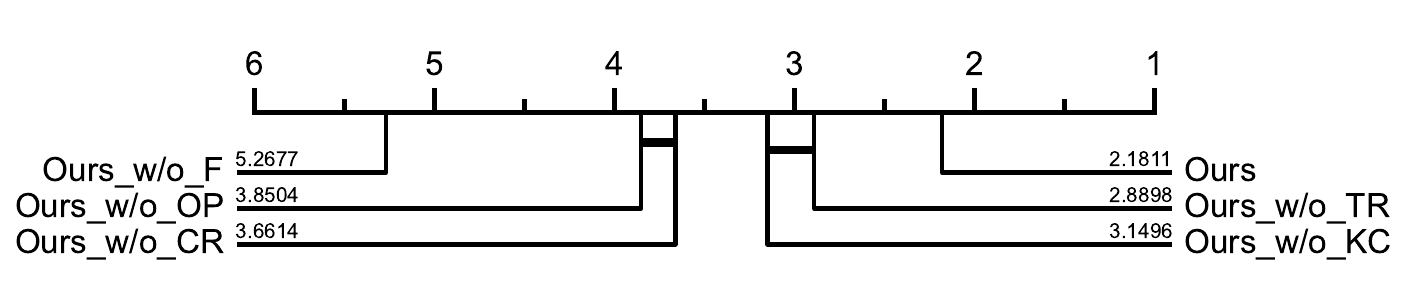}
    \label{subfig:cd_l}}\\

    \subfloat[AWCCI-VL]{\includegraphics[width=0.95\columnwidth]{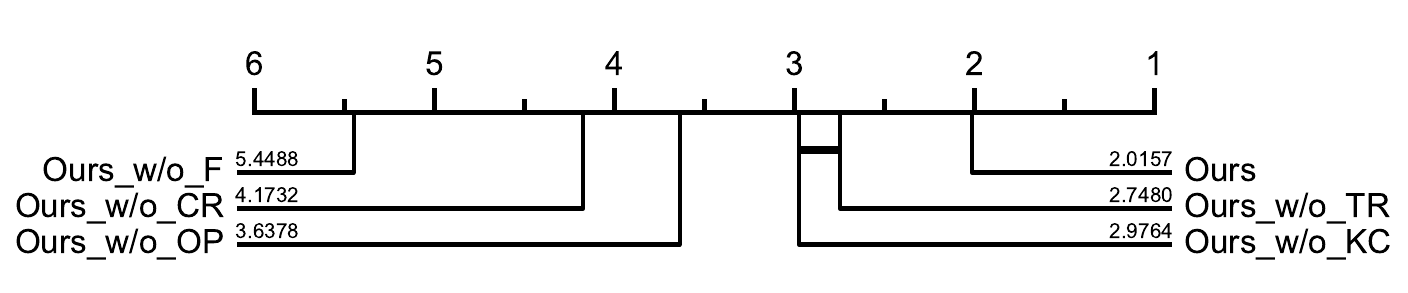}
    \label{subfig:cd_vl}}

    \caption{Ablation studies on the proposed designs.}
    \label{fig:CD_diagrams}
\end{figure}

We consider five ablation variants of \textbf{\texttt{MetaMTO}} corresponding to functional components in our proposed multi-role RL system: 1)~Ours\_w/o\_TR: we use a random~(uniformly) source task selector to replace the trained TR agent; 2)~Ours\_w/o\_KC: the portion of elite information to be transferred from the source sub-task to the target one no longer depends on the KC agent, instead, replaced by a random portion value sampled uniformly from $[0,1]$; 3)~Ours\_w/o\_OP, Ours\_w/o\_F and Ours\_w/o\_CR: these three variants represents ablation on the TSA agent group, where Ours\_w/o\_OP replaces the operator selection agent by a random selection mechanism, Ours\_w/o\_F and Ours\_w/o\_CR replace the corresponding configuration agents by fixed mutation strength $0.5$ and crossover rate $0.5$ respectively. To ensure an objective and comprehensive validation, the five models are trained on the five different AWCCI sets and then tested on the corresponding test sets. We leverage Wilcoxon-Holm post-hoc analysis~\cite{cd} for summarizing rank-based significance of all ablated variants and original \textbf{\texttt{MetaMTO}}~(denoted as Ours). The average rank is computed across testing results collected from 10 independent test runs. The corresponding CD diagrams are illustrated in Fig.~\ref{fig:CD_diagrams}, from which we can observe that: 
\begin{itemize}
    \item Consistent evidence can be attained in all five tests to support that all specific designs in our multi-role RL system make positive effects on the final performance of \textbf{\texttt{MetaMTO}} as a synergy, underscoring the possibility and importance of  meta-learning an EMT framework in a systematic and fine-grained fashion to control where, what and how to transfer knowledge;
    \item Ours v.s. Ours\_w/o\_TR: Despite the weak significance in AWCCI-VS, \textbf{\texttt{MetaMTO}} significantly outperforms the variant without task routing. This demonstrates that on more challenging MTO scenarios, selecting correct knowledge source for each sub-tasks is the first key step to ensure optimization effectiveness. The meta-trained TR agent in our \textbf{\texttt{MetaMTO}} could accurately locate similar task pairs, which facilitates subsequent EMT operations;
    \item Ours v.s. Ours\_w/o\_F: We found that a well-trained knowledge transfer strategy adaption policy might plays the most important role in boosting the underlying EMT framework. This also aligns with the existing research focus of MetaBBO: use RL for dynamic algorithm configuration.
\end{itemize} 

\section{Conclusion}\label{sec:6}
In this paper, we propose a novel MetaBBO framework termed as \textbf{\texttt{MetaMTO}} to address MTO problems. Our work presents several technical innovations to complement existing works for learning-assisted EMT. First, we develop a systematic and comprehensive MDP model to control the entire EMT process via explicit knowledge transfer. By formulating ``where, what and how'' to transfer knowledge as actionable algorithmic configuration, we propose a versatile multi-role RL system to achieve fine-grained and adaptive knowledge transfer. The multi-role system is equipped with attention-based architecture to facilitate both generic MTO modeling and inter-task similarity recognition. We train \textbf{\texttt{MetaMTO}} through PPO method and test the trained models in both in-distribution and out-of-distribution scenarios. Reproducible comparison results demonstrate that our approach consistently outperforms representative EMT approaches and advanced learning-assisted variants, while sharing similar computational overhead with these baselines. Further, the interpretability analysis reveals insightful patterns, including promising knowledge transfer rate, attention score heatmaps, and RL control preferences, which offer a reasonable explanation for the superiority of \textbf{\texttt{MetaMTO}}. Ablation studies on key designs in our approach not only validate their isolated effectiveness but also reveal that learning ``where, what and how'' to transfer knowledge in a holistic way is more promising than learning them separately.      


\bibliographystyle{IEEEtran}
\bibliography{refs}

\section*{Appendix}

\setcounter{section}{0}
\section{Technical details}
\subsection{Computational details of the state features}

In this section, we provide a description of the computational details in constructing the 5-dimensional feature vector that characterizes the optimization state information of each sub-task at every step of the evolutionary multitasking~(EMT) process.
To elaborate, the feature $s_1$ characterizes the diversity of a sub-task population in the decision space. Consider a population $X = \{x_1, x_2, ..., x_N\}$ of $N$ individuals for a given sub-task, where each individual is encoded in the unified space $[0,1]^D$. The computation of $s_1$ proceeds in two steps: fisrt, the standard deviation is calculated across the population for each dimension in the decision space, resulting in a $D$-dimensional vector of standard deviations; second, $s_1$ is obtained by taking the mean of this vector. This process can be mathematically formulated by the following equation:

\begin{equation}
    \sigma^{j}=\sqrt{\frac{1}{N}\sum_{i=1}^{N}(x_{i,j}-\mu_j)^2}
\end{equation}

\begin{equation}
    s_1 = \frac{1}{D}\sum_{j=1}^{D}\sigma^{j}
\end{equation}
where $\sigma_j$ represents the standard deviation of the $j$-th dimension across the population and $\mu_j$ is the mean value of the population along the $j$-th dimension. The second feature $s_2$ characterizes the convergence of the sub-task population in the objective space. Let $f$ denote the objective function of the sub-task, which is defined as a minimization problem. The feature $s_2$ is computed as follows. First, leveraging the known~(proxy) optimal value $f^*$ and the maximum objective value $f_{max}^0$ from the initial population as the baseline, the objective value of each individual is normalized to $[0,1]$. Then, $s_2$ is defined as the standard deviation of these normalized objective values. This process can be formally described by the following equations:

\begin{equation}
    \hat{f}(x_i)=\frac{f(x_i)-f^*}{f_{max}^0-f^*}
\end{equation}

\begin{equation}
    s_2=\sqrt{\frac{1}{N}\sum_{i=1}^{N}(\hat{f}(x_i)-\mu_{\hat{f}})^2}
\end{equation}
where $\hat{f}(x_i)$ is the normalized objective value of individual $x_i$ and $\mu_{\hat{f}}$ is the mean value of the normalized objective values. The third feature $s_3$ quantifies the degree of stagnation encountered during the optimization process. The stagnation situation is defined as the absence of improvement in the best-found objective value compared to the previous generation. $s_3=\frac{G_{stagnation}}{G}$ is then defined as the ratio of the cumulative stagnation count $G_{stagnation}$  to the total number of generations $G$. The fourth feature $s_4$ is a bool value that signals whether the best-so-far solution is updated. Let $f(X^t)$ and $f(X^{t+1})$ represent the best-so-far objective values at steps $t$ and $t+1$ respectively. Then, the mathematic formulation of the feature is $s_4 = \mathbb{I}(f(X^{t+1})<f(X^t))$, where $\mathbb{I}(\cdot)$ is the indicator function. Consequently, $s_4=1$ indicates a successful update of the best-so-far objective value. The fifth feature $s_5$ quantifies the effectiveness of knowledge transfer for a sub-task. Let $N_{tra}$ be the total number of transferred individuals of the sub-task and $N_{sur}$ be the number of these individuals that successfully survive the selection process. The feature $s_5 = \frac{N_{sur}}{N_{tra}}$ is defined as the survival rate of the transferred solutions. Finally, the complete state features of a sub-task is constructed as a vector containing the five features from $s_1$ to $s_5$, which holistically and succinctly encapsulates optimization information from many key aspects during the EMT process.

\subsection{Details of the problem set construction} 

The representative Multitask Optimization~(MTO) problem sets such as CEC2017\cite{CEC2017} and WCCI2020\cite{WCCI2020} are primarily limited by two issues: i) sparse and non-uniform distribution and ii) insufficient problem instances with various sub-task combinations. To address these issues, we propose a more comprehensive MTO problem set termed as AWCCI. In AWCCI, the sub-tasks within each MTO problem instance are generated by applying rotation and shifting operations to the following seven benchmark functions:

\begin{enumerate}
    \item Sphere:
        \begin{equation}
        \begin{split}
            f(\textbf{x})&=\sum_{i=1}^{D}z_i^2\\
            &\textbf{x}\in \left[-100,100\right]^D
        \end{split}
        \end{equation} \\
    \item Rosenbrock:
        \begin{equation}
        \begin{split}        f_2\left(\textbf{x}\right)&=\sum_{i=1}^{D-1}\left(100\left(z_i^2-z_{i+1}\right)^2+\left(z_i-1\right)^2\right)\\
        &\textbf{x}\in \left[-50,50\right]^D
        \end{split}
        \end{equation} \\
    \item Ackley:
        \begin{equation}
            \begin{split}       f_3(\textbf{x})&=-20exp\left(-0.2\sqrt{\frac{1}{D}\sum_{i=1}^{D}z_i^2}\right)\\
            &-exp\left(\frac{1}{D}\sum_{i=1}^{D}cos(2\pi z_i)\right)+20+e\\
            &\textbf{x}\in \left[-50,50\right]^D
            \end{split}
        \end{equation} \\
    \item Rastrigin:
        \begin{equation}
            \begin{split}           f_4(\textbf{x})&=\sum_{i=1}^{D}\left(z_i^2-10cos\left(2\pi z_i\right)+10\right)\\
            &\textbf{x}\in \left[-50,50\right]^D
            \end{split}
        \end{equation} \\
    \item Griewank:
        \begin{equation}
            \begin{split}
            f_5(\textbf{x})&=1+\frac{1}{4000}\sum_{i=1}^{D}z_i^2-\prod_{i=1}^{D}cos\left(\frac{z_i}{\sqrt{i}}\right)\\
            &\textbf{x}\in \left[-100,100\right]^D
            \end{split}
        \end{equation} \\
    \item Weierstrass:
        \begin{equation}
            \begin{split}    f_6(\textbf{x})&=\sum_{i=1}^{D}\left(\sum_{k=0}^{k_{max}}\left[a^kcos\left(2\pi b^k\left(z_i+0.5\right)\right)\right]\right)\\&-D\sum_{k=0}^{k_{max}}\left[a^kcos\left(2\pi b^k\cdot0.5\right)\right]\\
            &a=0.5,b=3,k_{max}=20\\
            &\textbf{x}\in \left[-0.5,0.5\right]^D
            \end{split}
        \end{equation} \\
    \item Schwefel:
        \begin{equation}
            \begin{split}    f_7(\textbf{x})&=418.9829\times D-\sum_{i=1}^{D}z_isin\left(|z_i|^{\frac{1}{2}}\right)\\
            &\textbf{x}\in \left[-500,500\right]^D
            \end{split}
        \end{equation}
\end{enumerate}
where for each benchmark function the dimension $D$ is 50 and $\textbf{z}=\textbf{w}^T\left(\textbf{x}-\textbf{s}\right)$. Note that $\textbf{w}$ denotes the random rotation matrix by applying Householder transformation\cite{refs:Householder_transformation} and $\textbf{s}$ denotes the shifting vector. The $lb$ and $ub$ denote the lower and upper bound of the shifting vector, which is within the same searching range as the decision space.

For the first issue, we apply five various distribution shifting levels $l\in\{0.05,0.1,0.2,0.3,0.4\}$ with incremental difficulties. As the configuration details presented in Table \ref{tab:AWCCI}, according to different shifting levels, the shifting vector of the sub-task is uniformly distributed within different bounded range. As a result, the AWCCI is further partitioned into five subsets, denoted respectively as  AWCCI-VS (with very small shifting), AWCCI-S (with small shifting), AWCCI-M (with median shifting), AWCCI-L (with large shifting) and AWCCI-VL (with very large shifting) corresponding to the specific shifting levels.

\begin{table}[t]
	\caption{Configuration of the AWCCI subset}	
    \centering
    \resizebox{0.9\columnwidth}{!}{
	\begin{tabular}{c c}
		\toprule[2pt]
		  Problem subset  & Shifting vector configuration(\textbf{s})\\
		\midrule 
		AWCCI-VS &  $\textbf{s} \sim 0.05\times(lb+U\left[0,1\right]^D\times\left(ub-lb\right))$  \\
        \specialrule{0em}{1pt}{1pt}
        AWCCI-S & $\textbf{s} \sim 0.1\times(lb+U\left[0,1\right]^D\times\left(ub-lb\right))$\\
        \specialrule{0em}{1pt}{1pt}
        AWCCI-M & $\textbf{s} \sim 0.2\times(lb+U\left[0,1\right]^D\times\left(ub-lb\right))$\\
        \specialrule{0em}{1pt}{1pt}
        AWCCI-L & $\textbf{s} \sim 0.3\times(lb+U\left[0,1\right]^D\times\left(ub-lb\right))$\\
        \specialrule{0em}{1pt}{1pt}
        AWCCI-VL & $\textbf{s} \sim 0.4\times(lb+U\left[0,1\right]^D\times\left(ub-lb\right))$\\
		\bottomrule [2pt]
	\end{tabular}
    }
    \label{tab:AWCCI}
\end{table}

For the second issue, let $\textbf{f}:=\{f_1,f_2,f_3,f_4,f_5,f_6,f_7\}$ be the benchmark function set, we begin by constructing $\sum_{i=1}^{7}C_7^i=127$ possible non-empty combinations of the seven benchmark functions in $\textbf{f}$. More specifically, each combination specifies the possible benchmark functions that each sub-task may adopt. Then for each combination, we generate a problem instance comprising 10 sub-tasks, with each sub-task randomly designated as one of the constitutive benchmark functions (with replacement) and instantiated by a randomly generated rotation matrix and shifting vector. Since the distribution of the shifting vector is associated with each subset, the generation process is repeated for all the five subsets respectively. Consequently, we generate 127 problem instances for each subset, resulting in a total of $127 \times 5 =635$ problem instances for the entire AWCCI problem set.

\end{document}